\documentclass{article}

\PassOptionsToPackage{numbers, compress}{natbib}

\usepackage[final]{neurips_2021}

\usepackage[utf8]{inputenc} 
\usepackage[T1]{fontenc}    
\usepackage{hyperref}       
\usepackage{url}            
\usepackage{booktabs}       
\usepackage{amsfonts}       
\usepackage{nicefrac}       
\usepackage{microtype}      
\usepackage{xcolor}         

\usepackage{graphicx}
\usepackage{subfigure}
\usepackage{multirow} 
\usepackage{amsmath}
\usepackage{amsthm}
\usepackage{subfigure}

\newcommand{\tsum}{\textstyle\sum}
\newcommand{\conv}{\textrm{Conv}}
\newcommand{\convo}{\textrm{Conv1}}
\newcommand{\convt}{\textrm{Conv3}}

\newenvironment{revised}{}{}

\title{Recurrence along Depth: Deep Convolutional Neural Networks with Recurrent Layer Aggregation}

\author{%
	Jingyu Zhao, Yanwen Fang and Guodong Li \\
	Department of Statistics and Actuarial Science \\
	The University of Hong Kong \\
	\texttt{\{gladys17, u3545683\}@connect.hku.hk, gdli@hku.hk}
}

\begin{document}

\maketitle

\begin{abstract}
This paper introduces a concept of layer aggregation to describe how information from previous layers can be reused to better extract features at the current layer. 
While DenseNet is a typical example of the layer aggregation mechanism, its redundancy has been commonly criticized in the literature. 
This motivates us to propose a very light-weighted module, called recurrent layer aggregation (RLA), by making use of the sequential structure of layers in a deep CNN. 
Our RLA module is compatible with many mainstream deep CNNs, including ResNets, Xception and MobileNetV2, and its effectiveness is verified by our extensive experiments on image classification, object detection and instance segmentation tasks. 
Specifically, improvements can be uniformly observed on CIFAR, ImageNet and MS COCO datasets, and the corresponding RLA-Nets can surprisingly boost the performances by 2-3\% on the object detection task. 
This evidences the power of our RLA module in helping main CNNs better learn structural information in images.
\end{abstract}

\section{Introduction} \label{sec:intro}
Convolutional neural networks (CNNs) have achieved notable success in computer vision tasks, crediting to their ability to extract high-level features from input images. 
Due to rapid growth in the depth of CNNs in recent years, the problem of how to pass information efficiently through layers often arises when designing deep architectures.
Residual connections \cite{he2016deep, he2016identity}, or skip connections, are now cornerstone components that act as information pathways to give layers direct access to previous ones and make training feasible with hundreds of layers.
In a deep feature hierarchy, higher-level features learned by the network are built upon simpler ones, but not necessarily on the layer right before it \cite{greff2017highway}. 
This conjecture can be supported by the stochastic depth training method \cite{huang2016deep}, where layers randomly gain access to previous ones.
Furthermore, it is discovered that entire layers can be removed from ResNets without impacting the performance.
This observation incubated the dense connectivity in DenseNets \cite{huang2017densely, huang2019convolutional}, where layers in a stage have direct access to all previous layers through fully connected skip connections.

We introduce a concept of layer aggregation to systematically study network designs on feature reuse, and a typical example is DenseNet.
However, within a densely-connected stage, the number of network parameters grows quadratically with respect to the number of layers.
These highly redundant parameters may extract similar features multiple times \cite{chen2017dual}, and limit the number of channels to store new information. 
This motivates us to propose a very light-weighted recurrent layer aggregation (RLA) module with much fewer parameters.
Specifically, we use recurrent connections to replace dense connectivity and achieve a parameter count independent of the network depth. 
The RLA module preserves the layer aggregation functionality, and it can be added to existing CNNs for better feature extraction.

\begin{figure}[ht]
	\centering
	\includegraphics[width=0.95\linewidth]{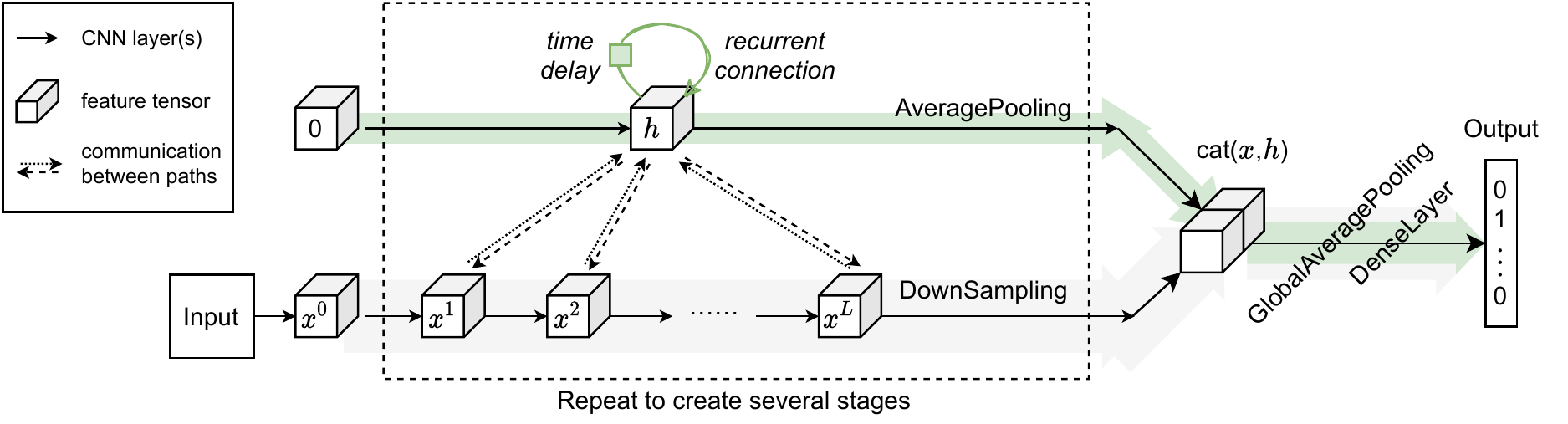}
	\caption{Schematic diagram of a CNN with recurrent layer aggregation for image classification.}
	\label{fig:schematic}
\end{figure}

Our RLA module is compatible with many CNNs used today. 
A schematic diagram is provided in Figure \ref{fig:schematic}, where $x$ represents hidden layers in a deep CNN and $h$ represents hidden states in RLA modules. 
\citet{lecun2015deep} pointed out that unfolded RNNs can be seen as very deep feedforward networks where all the hidden layers share the same weights.
When examined alone, an RLA module behaves similar to an unfolded RNN with layers of a deep CNN as its inputs.
But applying RLA modules to CNNs is more than a simple combination of RNN and CNN (e.g., \cite{moniz2016convolutional}).
As the key idea behind layer aggregation is to provide the current layer with a version of layer history, information exchange between main CNN and RLA modules is a must.
This results in a connection from the hidden unit of the RNN back to its inputs, which is hardly observable in other deep architectures.
Moreover, we do not use global average pooling when passing information from the main CNN to RLA, so that the historical information is enriched and contains spatial information. 

Despite its recurrent design, a convolutional-RNN-based RLA module can be easily implemented using standard convolutions with parameter sharing. 
Empirically, RLA modules are computationally light-weighted and impose only a slight increase in model parameters and computational cost. 
In terms of functionality, it serves purposes beyond channel attention and can be applied on top of channel attention modules.
We perform extensive experiments across a variety of tasks using different deep CNN architectures, including ResNets \cite{he2016deep, he2016identity}, ECANets \cite{wang2020eca}, Xception \cite{chollet2017xception} and MobileNetV2 \cite{sandler2018mobilenetv2}.
Our experimental results show that RLA consistently improves model performances on CIFAR, ImageNet and MS COCO datasets. 
On MS COCO, our RLA module can achieve 2-3\% gains in terms of average precision, which significantly outperforms other state-of-the-art networks with remarkable performances. 

The main contributions of our work are below:
(1) We introduce a definition of layer aggregation for analyzing CNN architectures.
(2) We propose a novel recurrent layer aggregation module, with motivation from the layer aggregation mechanism and time series analysis.
(3) We investigate detailed RLA module designs through ablation study and provide guidelines on applying RLA modules.
(4) We show the effectiveness of RLA across a broad range of tasks on benchmark datasets using several popular deep CNN architectures.

\section{Related work} \label{sec:related-work}

\paragraph{CNN-RNN hybrid models}
Deep CNNs with RLA modules should be easily distinguishable from other CNN-RNN hybrid networks, as RLA features information exchange between CNN and RNN at each layer and the corresponding recurrent update. 
In visual description tasks, RNNs are adopted to process the outputs of CNNs for sequence generation \cite{donahue2015long, vinyals2015show, shi2016end, wang2016cnn, chen2017sca}.
For video super-resolution, convolutional LSTMs can be used first to capture the temporal dependence between frames and then pass outputs to CNNs \cite{lim2017deep}.
There are only a few papers that apply the concept of recurrence to CNNs.
In RecResNet \cite{batsos2018recresnet}, a whole CNN is used recurrently for disparity map refinement. 
Moreover, in \cite{liang2015recurrent}, convolutional layers are used recurrently to produce "deep" CNNs before the debut of ResNets.

\paragraph{CNNs with branching paths}
CNNs with our RLA module have similar structures compared to those with branching paths, e.g., Inception networks \cite{szegedy2015going, ioffe2015batch, szegedy2016rethinking, szegedy2017inception}, or networks with grouped convolutions, e.g., AlexNet \cite{krizhevsky2012imagenet} and ResNeXt \cite{xie2017aggregated}. 
The main difference is that our RLA module has shared convolutional kernels along depth. 
When RLA is applied to ResNets, the resulting networks can be viewed as compressed Dual Path Networks \cite{chen2017dual}.

\paragraph{Attention mechanisms in CNNs}
Attention mechanisms have been incorporated into CNNs mainly in the form of spatial attention and channel attention. 
Successful applications include
image captioning \cite{xu2015show, chen2017sca}, 
visual question answering \cite{zhu2016visual7w}, 
image classification \cite{wang2017residual, fu2019dual}, 
semantic segmentation \cite{li2018pyramid}, 
face verification \cite{chen2018mobilefacenets}, and 
video frame interpolation \cite{choi2020channel}.
An outstanding design of channel attention would be the Squeeze-and-Excitation (SE) block \cite{hu2018squeeze}, which is later adopted by MobileNetV3 \cite{howard2019searching} and EfficientNet \cite{tan2019efficientnet} and developed into compressed variants \cite{wang2020eca}. 
The proposed layer aggregation mechanism has a connection with channel attention across multiple layers, and meanwhile, it can be applied to networks with channel attention modules.

\paragraph{RNN variants}
Since RLA modules adopt the form of a convolutional RNN, RNN variants give rise to different RLA designs.
For example, the hidden feature maps in RLA can also be updated by the convolutional counterparts of LSTM \cite{hochreiter1997long, shi2015convolutional} or GRU cells\cite{cho2014learning}.
In terms of architectural connectivity, skip connections can be introduced to RNN hidden states \cite{lin1996learning, zhang2016architectural, chang2017dilated, dipietro2017analyzing}.
Moreover, for deep RNNs, different layers of state variables can be updated at different frequencies \cite{el1996hierarchical}. 
Correspondingly, we may derive deep RLA modules that formulate stage aggregation on top of layer aggregation.

\section{Recurrent layer aggregation modules} \label{sec:RLA_module}
This section first introduces a concept of layer aggregation and some parsimonious models in time series analysis, which together motivate a type of light-weighted recurrent layer aggregation (RLA) modules by making use of the sequential structures of deep CNNs.

\subsection{Layer aggregation} \label{sec:LA}
Consider a deep CNN with $ x^t $ being the hidden features at the $t$th layer and $x^0$ being the input, where $L$ is the number of layers, and $1\leq t\leq L$.
The CNN is said to have the \textit{layer aggregation mechanism} if
\begin{equation}\label{eq:layer-agg}
	A^t  = g^t(x^{t-1},x^{t-2},\ldots,x^0) \hspace{5mm}\text{and}\hspace{5mm}
	x^t  = f^t (A^{t-1}, x^{t-1}), 
\end{equation}
where $ f^t $ denotes the transformation that produces new feature map at layer $t$ of the CNN, $ A^t $ represents the aggregated information up to $(t-1)$th layer, and $g^t$ is the corresponding transformation that summarizes these layers.
For CNNs without skip connections, according to our definition, they do not involve layer aggregation since $x^t  = f^t (x^{t-1})$, i.e., the information in $x^l $s with $ l < t-1 $ is only available to $ x^{t} $ through $x^{t-1}$.
On the other hand, the Hierarchical Layer Aggregation \cite{yu2018deep} can be shown to have such mechanism since it satisfies Eq. \eqref{eq:layer-agg}.

It is usually more convenient to consider an additive form for the summarizing transformation, i.e. 
\begin{equation}\label{eq:layer-agg1}
	A^t  =  \tsum_{l=0}^{t-1} g_{l}^t (x^l),
\end{equation}
where an additive constant $c$ may also be included, and $ g_{l}^t $ can be one or more convolutions, e.g., a set of 1x1-3x3-1x1 convolutions that compose a bottleneck block.
Consider a DenseNet layer
\begin{equation} \label{eq:densenet}
	x^t = \convt^t \left[\convo^t (\text{Concat}(x^0, x^1, ..., x^{t-1})) \right] ,
\end{equation}
where $\convo^t$ and $ \convt^t $ denote 1x1 and 3x3 convolutions, respectively. 
It can be verified that
\begin{equation}\label{eq:dense-agg}
	\convo^t (\text{Concat}(x^0, x^1, ..., x^{t-1})) = \tsum_{l=0}^{t-1} \convo_{l}^t (x^l),
\end{equation}
where the kernel weights of $ \convo_{l}^t $ form a partition of the weights in $ \convo^t $. As a result,
\begin{equation*}
	A^t = \tsum_{l=0}^{t-1} \convo_{l}^{t+1} (x^l) \hspace{5mm}\text{and}\hspace{5mm}
	x^t  = \convt^t \left[ A^{t-1}+\convo_{t-1}^t (x^{t-1}) \right],
\end{equation*}
i.e., it is a typical example of the layer aggregation mechanism.

We next consider the pre-activated ResNets \cite{he2016identity}. 
On the one hand, according to the ResNet update 
$ x^t = x^{t-1} + g_{t-1} (x^{t-1}) $, it may be concluded to involve no layer aggregation.
On the other hand, by recursively applying the update, we have $ x^t = \sum_{l=0}^{t-1} g_{l} (x^l) + x^0 $, which satisfies Eq. (\ref{eq:layer-agg}) since $ A^t =\sum_{l=0}^{t-1} g_{l} (x^l) + x^0 $ and $x^t=A^{t-1} + g_{t-1} (x^{t-1})$. 
It is noticeable that $ g_{l} $ depends only on $l$, the ordinal number of previous layers, and we will discuss such a pattern in the next subsection.

\paragraph{Connection with channel attention}
The layer aggregation mechanism in the original DenseNet can be interpreted as channel attention across layers. 
Specifically, the \textit{multi-head masked self-attention} \cite{vaswani2017attention} mechanism can be written as
\begin{equation} \label{eq:multi-head}
	\text{Attention}^t
	= \text{Concat}(\text{head}_1^t, ..., \text{head}_m^t) \text{ with head}_i^t 
	= \tsum_{l=0}^{t-1} s_{l, i}^{t} \, x^{l} , \text{ for } i = 1, ..., m, 
\end{equation}
where query, key and value are all set to be $x^t$, $ s_{l,i}^t $s are scalar similarity measures and $m$ is the number of attention heads. 

Note that, when considering DenseNet layers, feature maps are tensors instead of matrices.
Suppose that all $x^l$'s have the same number of channels, denoted by $k$, and then $ \convo^t $ has $ tk $ input and $ m $ output channels.
Denote the $ tk \times 1\times 1\times m$ kernel weights of $\convo^t$ by $ \{W_{j, i}^{t}\}$.
Let $\text{head}_i^t$ be the $i$th channel in $ \text{Attention}^t $, and
$ x_c^{l} $ be the $c$th channel of the feature tensor $ x^l $,
then it holds that $ \text{head}_i^t = \sum_{l, c} W_{j(l, c), i}^{t} \, x_c^{l} , $ 
where $ j(l, c) = lk + c $ is an index. 
As a result, DenseNet updates coincide with a natural extension of Eq. (\ref{eq:multi-head}) to tensor variables.

However, the above attention mechanism is too meticulous in a CNN context.
For example, it will collapse when simply changing $ \convo^t $ in DenseNets to a 3x3 convolution. 
In contrast, the layer aggregation interpretation of DenseNet is robust to such a modification.

\begin{revised}
	Our purpose of the discussions above is two-fold:
	(1) we build the relationship that layer aggregation generalizes channel attention across layers based on the DenseNet architecture; and thus, 
	(2) layer aggregation is not a substitute to channel attention modules, but complementary to them, which inspires us to add the proposed RLA modules to ECAnets in Section \ref{sec:exp}.
\end{revised}

\subsection{Sequential structure of layers in deep networks} \label{sec:SS}
The complexity of summarizing transformations can drop dramatically from \eqref{eq:layer-agg} to \eqref{eq:layer-agg1}, while it may still be complicated.
Actually DenseNets have been frequently criticized for their redundancy.
Note that the layers $\{x^t\}$ form a sequential structure, and hence this subsection will introduce some related concepts in time series analysis such that we can further compress the summarizing transformations at \eqref{eq:layer-agg} and \eqref{eq:layer-agg1}. 
\begin{revised}
	More background on time series is available in the supplementary material A.2.
\end{revised}

We first simplify Eq. \eqref{eq:layer-agg1} by assuming that $g_l^t$ depends on the lag $t-l$ only, and this is a typical setting in time series analysis.
If all $x^t$s are scalars and all related transformations are linear, we have
\begin{equation} \label{eq:AR}
	x^t = \beta_{1}x^{t-1}  + ... + \beta_{t-1}x^{1}  +  \beta_{t} x^{0} = \tsum_{l=0}^{t-1} \beta_{t-l} x^{l}.
\end{equation} 
It can also be rewritten into $\tsum_{l=1}^{\infty} \beta_{l} x^{t-l}$ with $x^s=0$ for $s<0$,
which has a form of autoregressive (AR) models in time series, $x^t=\tsum_{l=1}^{\infty} \beta_{l} x^{t-l} +i^t$, with $i^t$ being the additive error.

In time series analysis, the above AR model, which is usually referred to as the AR$(\infty)$ model, is mainly for theoretical discussions, and the commonly used running model is the autoregressive moving average (ARMA) model since it has a parsimonious structure.
Consider a simple ARMA$(1,1)$ model, $x^t = \beta x^{t-1}  + i^t - \gamma i^{t-1} $.
It can be verified to have an AR$(\infty)$ form of 
\begin{align}
	\begin{split}\label{eq:arma}
	x^t&=\beta x^{t-1}  + i^t - \gamma i^{t-1} = \beta x^{t-1} - \gamma (x^{t-1} - \beta x^{t-2}) + i^t - \gamma^2 i^{t-2} \\
	 &=\cdots =\tsum_{l=1}^{\infty} (\beta-\gamma)\gamma^{l-1} x^{t-l} +i^t,
	\end{split}
\end{align}
where equation $i^s=x^s -\beta x^{s-1}+\gamma i^{s-1}$ is recursively applied for all $s\leq t-1$, 
while there are only two parameters involved.
This motivates us to use a similar pattern to construct a more parsimonious form for Eq. \eqref{eq:AR}.
Let 
\begin{equation}\label{eq:rnn}
	h^t=\alpha x^{t-1}+\gamma h^{t-1}\hspace{5mm}\text{and}\hspace{5mm}
	x^t=\beta_1x^{t-1}+\beta_2h^{t-1}.
\end{equation}
Assuming $ h^0 = 0 $, we then have $h^t=\sum_{l=1}^t\alpha \gamma^{l-1}x^{t-l}$ and $x^t=\beta_1x^{t-1}+\beta_2\alpha \sum_{l=1}^{t-1}\gamma^{l-1}x^{t-1-l}$, which has a form of \eqref{eq:AR} but involves at most four parameters.
It is worthy to point out that $h^t$ plays a role similar to that of $i^t$, and Eq. \eqref{eq:rnn} is similar to the principle of RNNs \cite{connor1994recurrent}.

We may also consider the case that $g_l^t$ at \eqref{eq:layer-agg1} depends on $l$ only, i.e. $x^t  = f^t (\tsum_{l=0}^{t-2} g_{l} (x^l), x^{t-1})$, which leads to a ResNet-type layer.
From \eqref{eq:rnn}, for $s>0$, the information in $x^{t-s}$ will be reused when extracting features $x^t$ by an amount that depends on lag $s$ only and decays at a rate of $\gamma^s$.
However, as mentioned in Section 3.1, ResNet can be viewed as layer aggregation in a different pattern, and the amount of information in $x^l$ reused when extracting features $x^t$ depends on the ordinal number $l$ only.
We next conduct a small experiment to compare the ResNet-type simplification with that at \eqref{eq:AR}.

\paragraph{Comparison of two simplifications}
The experiment is carried out based on the $100$-layer DenseNet-BC ($k=12$) \cite{huang2017densely}.
According to the two simplifications, we modify the DenseNet update \eqref{eq:densenet} into the following versions:
\begin{align} 
\mbox{(Shared-Lag)} & \quad x^t = \convt^t (\convo_{1} (x^{t-1}) + \cdots + \convo_{t-1} (x^{1}) + \convo_{0}^t (x^{0})), \label{eq:densenet-arp} \\
\mbox{(Shared-Ordinal)} & \quad x^t = \convt^t (\convo_{t-1} (x^{t-1}) + \cdots + \convo_{1} (x^{1}) + \convo_{0}^t (x^{0})), \label{eq:densenet-ord}
\end{align}
which correspond to \eqref{eq:AR} and ResNet, respectively.
The above formulations require $ \convo_l $ to be compatible with all previous layers.
For DenseNet-BC ($k=12$), the input to dense layers $ x^0 $ has $ 3 + 2\times k = 27 $ channels, while all later layers $ x^l, 1 \leq l \leq 16 $, have $k=12$ channels.
Thus, we keep the convolutions applied to $x^0$ unchanged and only modify the other 1x1 convolutions.
Detailed explanations of the network structure of \eqref{eq:densenet-arp} are provided in the supplementary material A.2.

\begin{table}[t]
	\caption{Classification error of DenseNet and its parameter sharing variants on the CIFAR-10 dataset.} 
	\centering
	\small
	\begin{tabular}{lcc}
		\toprule
		Model        	& Params & Error (\%) \\
		\midrule
		DenseNet     	& 0.80M	   & 5.80$\pm$ 0.33 \\
		- Shared-Lag 	& 0.60M    & 6.22 \\
		- Shared-Ordinal& 0.60M    & 6.29 \\
		\bottomrule
	\end{tabular}
	\label{tab:shared}
	\vskip -0.1in
\end{table}

Table \ref{tab:shared} presents the comparison of parameter count and accuracy on CIFAR-10. 
Detailed training settings are the same as other experiments on CIFAR-10 in Section \ref{sec:exp}.
Sharing the 1x1 convolutions reduces the number of parameters by 25\% (0.20M), while the accuracy only drops by about 0.5\%.
This observation is consistent with the parameter sharing experiment by \cite{liao2016bridging}, where they found that a ResNet with shared weights retains most of its performance on CIFAR-10.
And the better performance of Shared-Lag encourages us to pursue a more parsimonious structure along the direction of \eqref{eq:AR}.

Furthermore, we extract the weights learned by the shared 1x1 convolutions and produce Figure \ref{fig:shared1x1weights}, following Figure 5 in \cite{huang2017densely}.
A first observation is that the strongest signal in Shared-Lag (lag=1) is almost twice as strong as that in Shared-Ordinal (layer=1).
Secondly, we spot a quick decaying pattern in the plot for Shared-Lag, which is similar to the \textit{exponential decay} pattern in an ARMA model or an RNN \cite{zhao2020rnn}.
This observation provides us empirical evidence to adopt the principle at \eqref{eq:rnn} to design a light-weighted layer aggregation mechanism.

\begin{figure}[!htp]
\begin{minipage}[t]{0.5\linewidth}
	\centering
	\includegraphics[width=1.1\linewidth]{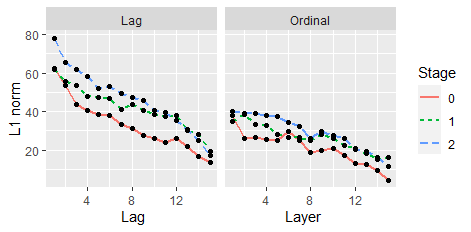}
	\caption{$L_1$ norm of the shared weights.
	\label{fig:shared1x1weights}}
\end{minipage}
\hfill
\begin{minipage}[t]{0.45\linewidth}
	\centering
	\includegraphics[width=0.75\linewidth]{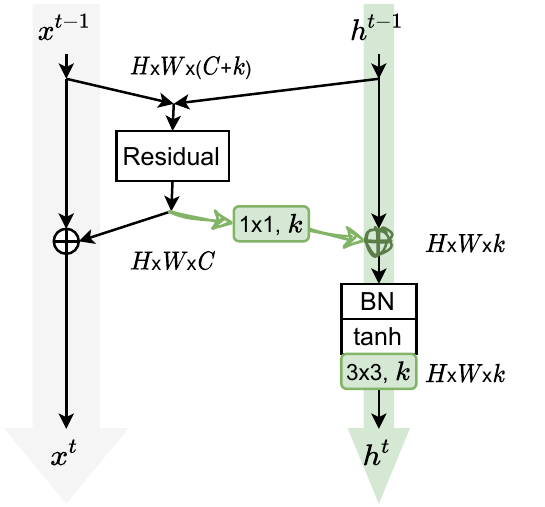}
	\caption{Diagram of an RLA module.
	\label{fig:RLA}}
\end{minipage}
\end{figure}

\subsection{Recurrent layer aggregation} \label{sec:RLA}
By adopting the idea at \eqref{eq:rnn}, we introduce a recurrent layer aggregation (RLA) mechanism below,
\begin{equation}\label{eq:RLA}
	h^t  = g^t (h^{t-1}, x^{t-1})\hspace{5mm}\text{and}\hspace{5mm}
	x^t=f^t (h^{t-1}, x^{t-1}),
\end{equation}
where $ A^t =h^t$ represents the recurrently aggregated information up to $(t-1)$th layer, and it can be shown to have an additive form of \eqref{eq:layer-agg1} under a linearity assumption; see supplementary material A.3.
By considering different forms of $g^t$, we can achieve various \textit{RLA modules} to help the main CNN better extract features, and weight sharing can be further used to regularize the layer aggregation. 
The form of $f^t$ is mainly determined by the currently used mainstream CNNs, and the combination of aggregated information may be implemented in different ways.

Figure \ref{fig:RLA} gives an example for a CNN block with a residual connection, and \begin{revised}legends are provided at the bottom of Figure \ref{fig:ablation}\end{revised}.
Its update can be formulated as 
\begin{equation*}
	h^t=g_2[g_1(y^t)+h^{t-1}]\hspace{5mm}\text{and}\hspace{5mm}
	x^t=y^t+x^{t-1},
\end{equation*}
where $y^t=f_1^t [\text{Concat}(h^{t-1}, x^{t-1})]$.
Specifically, inputs $x^{t-1}$ and $h^{t-1}$ are combined via concatenation and then passed to a residual unit to extract new features.
One copy of this new feature map is joined with the residual connection, and the other copy is compressed via the shared 1x1 convolution and then added to $h^{t-1}$.
Then, the RLA path goes through a pre-activated ConvRNN unit comprised of \begin{revised}an unshared\end{revised} batch normalization, a tanh activation, and a recurrent 3x3 convolution. 

Practically, backbone CNNs often comprise of several stages, and RLA modules can cater to various stage structures following the guidelines below:
\begin{enumerate}
	\item Group main CNN layers/blocks into stages so that feature maps within the same stage have the same resolution, e.g., ResNet-50/101/152 has 4 stages and MobileNetV2 has 5 stages.
	\item For each stage, define (a) a shared recurrent unit, allowing information to be compressed differently at various stages, and (b) a shared 1x1 convolution, to change the channels of the residual feature maps for adding it to the RLA hidden state.
	\item (\textit{RLA path}) RLA hidden state $h^0$ is initialized as zero. Before entering the next stage, the RLA hidden states are downsampled to the correct resolution via average pooling.
	\item (\textit{RLA output}) For classification, the last hidden state is concatenated with the main CNN output and passed to the classifier. For detection and segmentation, RLA output can be discarded.
\end{enumerate}

\section{Experiments} \label{sec:exp}
We verify the effectiveness of our RLA module in Figure \ref{fig:RLA} on image classification, object detection and instance segmentation tasks using CIFAR, ImageNet and MS COCO datasets. 
All experiments are implemented on four Tesla V100 GPUs.
\footnote{\revised{Our implementation and weights are available at \url{https://github.com/fangyanwen1106/RLANet}.}}
More implementation details and tabular descriptions of network architectures are provided in Section B of the supplementary material.

\begin{table*}[t]
	\caption{Classification error (\%) on CIFAR-10 and CIFAR-100 test sets. Results are highlighted in bold if RLA-Net outperforms our re-implemented baseline model shown in the column "Re-impl.".}
	\vskip -0.1in
	\label{tab:cifar}
	\begin{center}
		\begin{small}
			\begin{tabular}{lcrrrrrrr}
				\toprule
				Dataset & Model  &  Params  & FLOPs  &  Original & Re-impl. &  RLA-Net & $\Delta$ Params  &  $\Delta$FLOPs  \\
				\midrule
				\multicolumn{1}{c}{\multirow{3}{*}{C-10}} & ResNet-110  &  1.73M  &  8.67M  &  6.37  &  6.35  &  \bf{5.88} &  +0.07M  &  +0.37M  \\
				\multicolumn{1}{c}{}  &  ResNet-164  &  1.72M  &  8.55M  &  5.46  &  5.72  &  \bf{4.95} & +0.02M  &  +0.19M  \\
				\multicolumn{1}{c}{}  &  Xception  &  2.70M  &  13.54M  &  -  &  6.03  &  \bf{5.78} &  +0.07M & +0.33M \\
				\midrule
				\multicolumn{1}{c}{\multirow{3}{*}{C-100}} & ResNet-110  &  1.74M  &  8.69M  &  26.88  &  28.51  &  \bf{27.44} &  +0.07M  &  +0.38M  \\
				\multicolumn{1}{c}{}  & ResNet-164  &  1.74M  &  8.67M  &  24.33  &  25.22  &  \bf{23.78} &  +0.02M  &  +0.19M  \\
				\multicolumn{1}{c}{}  & Xception  &  2.75M  &  13.77M  &  -  &  24.31  &  \bf{24.00} &  +0.06M  &  +0.33M  \\
				\bottomrule
			\end{tabular}
		\end{small}
	\end{center}
	\vskip -0.1in
\end{table*}

\subsection{CIFAR and analysis}
The two CIFAR datasets, CIFAR-10 and CIFAR-100, consist of 60k $32 \times 32$ RGB images of 10 and 100 classes. 
The training and test sets contain 50k and 10k images. 
We adopt a train/validation split of 45k/5k and follow the widely used data preprocessing scheme in~\cite{he2016deep, huang2017densely}. 
All the networks are trained from scratch using Nesterov SGD with momentum of 0.9, $l_2$ weight decay of $10^{-4}$, and a batch size of 128 for 300 epochs.
The initial learning rate is set to 0.1 and is divided by 10 at 150 and 225 epochs.
We choose the model with the highest validation accuracy.

We report the performance of each baseline model and its RLA-Net counterpart on CIFAR-10 and CIFAR-100 test sets in Table \ref{tab:cifar}. 
We observe that RLA-Net consistently improves performance across various networks with different depths, blocks and patterns of stages. 
For the three models in Table \ref{tab:cifar}, adding the RLA module introduces about a 1\%-4\% increase in parameters and FLOPs. 
Meanwhile, our RLA-Net improves the accuracies of ResNet-110 and ResNet-164 by about 0.5\% and 1.0\% on CIFAR-10 and CIFAR-100, respectively.


\begin{table}[t] 
	\vskip -0.1in
	\caption{Comparisons of single-crop error on the ILSVRC 2012 validation set with center crop of size $224 \times 224$. All models were trained with a crop size of $224 \times 224$.
	The last column report increases in Top-1 accuracy by adding RLA.
	$ \dagger$ Trained with some training tricks and architectural refinements in \cite{he2019bag}; see Section B2.2.}
	\vskip -0.1in
	\label{tab:imagenet}
	\begin{center}
		\begin{small}
			\begin{tabular}{l|cc|ccc}
				\toprule
				Baseline+Module  &  Params  &  FLOPs  & Top-1 err. &  Top-5 err. & $\uparrow$Acc1\\
				\midrule
				ResNet-50 \cite{resnetgit} &  24.37M  &  3.83G  &  24.70  &  7.80  &  \\   
				+RLA (Ours) &  24.67M  &  4.17G  &  22.83  &  6.58  &  1.87  \\
				+SE \cite{hu2018squeeze} &   26.77M  &  3.84G  &  23.29  &  6.62  &    \\
				+CBAM \cite{woo2018cbam} &  26.77M  &  3.84G  &  22.66  &  6.31  &  \\
				+ECA \cite{wang2020eca} &   24.37M  &  3.83G  &  22.52  &  6.32  &    \\
				+ECA+RLA (Ours) &  24.67M &  4.18G  &  {\color{blue}\textbf{22.15}}  &  {\color{blue}\textbf{6.11}}  &  0.37  \\
				\midrule
				ResNet-101 \cite{resnetgit} &  42.49M  &  7.30G  &  23.60 &  7.10  &   \\
				+RLA (Ours) &  42.92M  &  7.79G  &  21.48  &  5.80  &  2.12  \\
				+SE \cite{hu2018squeeze} &  47.01M  &  7.31G  &  22.38  &  6.07  &   \\
				+CBAM \cite{woo2018cbam} &  47.01M  &  7.31G  &  21.51  &  5.67  &  \\
				+ECA \cite{wang2020eca} &  42.49M  &  7.30G  &  21.35  &  5.66  &   \\
				+ECA+RLA (Ours) & 42.92M  &  7.80G  &  {\color{blue}\textbf{21.13}}  &  {\color{blue}\textbf{5.61}}  &  0.22  \\
				\midrule
				ResNet-152 \cite{resnetgit} &  57.40M  &  10.77G  &  23.00 &  6.70  &   \\
				+RLA (Ours) & 57.96M  &  11.47G  &  21.22  &  5.65  &  1.78  \\
				+SE \cite{hu2018squeeze} &  63.68M  &  10.78G  &  21.57  &  5.73  &   \\
				+ECA \cite{wang2020eca} &  57.41M  &  10.78G  &  21.08  &  {\color{blue}\textbf{5.45}}  &   \\
				+ECA+RLA (Ours) &  57.96M  &  11.48G  &  {\color{blue}\textbf{20.66}}  &  5.49  &  0.42  \\
				\midrule
				MobileNetV2 \cite{wang2020eca, sandler2018mobilenetv2} &  3.34M  &  299.6M  &  28.36 &  9.80  &    \\
				+RLA (Ours) &  3.46M  &  351.8M  &  27.68  &  \textbf{9.18}  &  0.68  \\
				+SE \cite{hu2018squeeze} &  3.51M  &  300.3M  &  27.58  &  9.33  &   \\
				+ECA \cite{wang2020eca} &  3.34M  &  300.1M  &  27.44  &  9.19  &    \\
				+ECA+RLA (Ours) &  3.46M  &  352.4M  &  {\color{blue}\textbf{27.07}}  &  {\color{blue}\textbf{8.89}}  & 0.37   \\
				\midrule
				DenseNet-161 ($k=48$) \cite{wang2020eca} &  27.35M  &  7.34G  &  22.35 &  6.20  &    \\
				DenseNet-264 ($k=32$) \cite{wang2020eca} &  31.79M  &  5.52G  &  22.15 &  6.22  &
				   \\
				ResNet-200 \cite{wang2020eca} &  74.45M  &  14.10G  &  21.80  &  6.00  &  \\
				DPN-68 \cite{yang2020gated} & 12.80M & 2.50G & 23.60 & 6.90 & \\
				DPN-92 \cite{yang2020gated} & 38.00M & 6.50G & 20.70 & 5.40 & \\
				DPN-98 \cite{yang2020gated} & 61.60M & 11.70G & 20.20 & 5.20 & \\
				AOGNet-12M \cite{yang2020gated} & 11.90M & 2.40G & 22.30 & 6.10 & \\
				AOGNet-40M \cite{yang2020gated} & 40.30M & 8.90G & 19.80 & 4.90 & \\
				HCGNet-B \cite{yang2020gated} & 12.90M & 2.00G & 21.50 & 5.90 & \\
				HCGNet-C \cite{yang2020gated} & 42.20M & 7.10G & 19.50 & 4.80 & \\
				RLA-ResNet-50 (Ours) $ \dagger$ &  24.67M  &  4.17G  &  {{20.25}}  &  {{5.12}}  &  \\
				RLA-ECANet-50 (Ours) $ \dagger$ &  24.67M  &  4.18G  &  {\color{blue}\textbf{19.31}}  &  {\color{blue}\textbf{4.71}}  &  \\
				\bottomrule
			\end{tabular}
		\end{small}
	\end{center}
	\vskip -0.1in
\end{table}

\subsection{ImageNet classification}
This subsection reports experiments on the ImageNet LSVRC 2012 dataset \cite{deng2009imagenet}.
We employ our RLA module on the widely used ResNet \cite{he2016deep} architectures and the light-weighted MobileNetV2 \cite{sandler2018mobilenetv2}, as well as their counterparts with the state-of-the-art channel attention ECA modules \cite{wang2020eca}.
For MobileNetV2, depthwise separable convolutions are used for the shared 3x3 convolutions \begin{revised}to adapt to the inverted residual blocks of MobileNetV2.\end{revised}
When both RLA and ECA modules present, ECA modules are placed according to \cite{wang2020eca}, and the output of the ECA module is then passed to our RLA module through the shared 1x1 convolution.
We compare our RLA-Nets with several state-of-the-art attention-based CNN architectures, including SENet \cite{hu2018squeeze}, CBAM \cite{woo2018cbam} and ECANet \cite{wang2020eca}.

For training RLA-ResNets, we follow the same data augmentation and hyper-parameter settings as in \cite{he2016deep, huang2017densely}. 
All the networks are trained using SGD with momentum 0.9, $l_2$ weight decay of $10^{-4}$, and a mini-batch size of \begin{revised}256\end{revised} on 4 GPUs.
We train models for 120 epochs from scratch and use the weight initialization strategy described in \cite{he2015delving}.
The initial learning rate is set to 0.1 and decreased by a factor of 10 every 30 epochs.
For the light-weighted model MobileNetV2, we train the model on 2 GPUs within 400 epochs using SGD with weight decay of 4e-5, momentum of 0.9, and a mini-batch size of 96, following the settings in \cite{sandler2018mobilenetv2}. 
The initial learning rate is set to 0.045, decreased by a linear decay rate of 0.98.
All the parameters and FLOPs shown in the tables are computed by our devices.

Table \ref{tab:imagenet} shows that our RLA-Nets obtain better results compared with their original counterparts while introducing small extra costs.
Though RLA modules introduce more FLOPs when compared with channel attention modules, we find that the extra training time per epoch is almost the same.
Specifically, adding SE, ECA or RLA module to ResNet-101 costs about 15\%, 15\% or 19\% more training time.
RLA-ResNets introduce about 1\% increase in the number of parameters, leading to 1.9\%, 2.1\% and 1.8\% increases in top-1 accuracy for RLA-ResNet-50/101/152, respectively.
Similar results can be observed on the light-weighted MobileNetV2 architectures.
Furthermore, combinations of RLA and ECA modules yield remarkably better results, achieving the best performances among these models.
\begin{revised}These further improvements based on ECANets verify our perspective that the functionalities of channel attention modules and our proposed RLA module are complementary.\end{revised}


\begin{revised}
	Last but not least, our model with some tricks in \cite{he2019bag} can perform remarkably better than many popular networks and state-of-the-art networks with delicate designs, including DPN \cite{chen2017dual}, AOGNet \cite{li2019aognets} and HCGNet \cite{yang2020gated}, which are designed to combine the advantages of ResNet and DenseNet.
	The notable improvements demonstrate that any tricks or refinements used to further improve the original ResNets can be directly applied to our model to further improve the model performance.
	And RLA modules can be treated as one of the beneficial architectural refinements to be adopted by state-of-the-art CNNs as well.
\end{revised}

\subsection{Object detection and instance segmentation on MS COCO}

\begin{table}[t] 
	\caption{Object detection results of different methods on COCO val2017.}
	\vskip -0.1in
	\label{tab:coco_detection}
	\begin{center}
		\begin{small}
			\begin{tabular}{l|c|ccc|ccc|c}
				\toprule
				Methods  & Detector & $AP$  &  $AP_{50}$  &  $AP_{75}$  &  $AP_S$  &  $AP_M$  &  $AP_L$ & $\uparrow AP$ \\
				\midrule
				ResNet-50 &  \multirow{10}{*}{Faster R-CNN}  & 36.4 & 58.2 & 39.2 & 21.8 & 40.0 & 46.2 &  \\
				+RLA (Ours) & &  \textbf{38.8} & 59.6 & \textbf{42.0} & 22.5 & \textbf{42.9} & \textbf{49.5} & 2.4 \\
				+SE &  & 37.7 & 60.1 & 40.9 & 22.9 & 41.9 & 48.2 &  \\
				+ECA &  &  38.0 & 60.6 & 40.9 & 23.4 & 42.1 & 48.0 &  \\
				+ECA+RLA (Ours) &  &  {\color{blue} \textbf{39.8}} & {\color{blue}\textbf{61.2}}  & {\color{blue} \textbf{43.2}} & {\color{blue} \textbf{23.9}} & {\color{blue} \textbf{43.7}} & {\color{blue} \textbf{50.8}} & 1.8 \\
				\cmidrule{1-1} \cmidrule{3-9}
				ResNet-101 &   & 38.7 & 60.6 & 41.9 & 22.7 & 43.2 & 50.4 &  \\
				+RLA (Ours) &  & \textbf{41.2} & 61.8 & \textbf{44.9} & 23.7 & \textbf{45.7} & \textbf{53.8} & 2.5 \\
				+SE &  & 39.6 & 62.0 & 43.1 & 23.7 & 44.0 & 51.4 &  \\
				+ECA &  & 40.3 & 62.9 & 44.0 & {\color{blue} \textbf{}}24.5 & 44.7 & 51.3 &  \\
				+ECA+RLA (Ours) &  & {\color{blue} \textbf{42.1}} & {\color{blue}\textbf{63.3}}  & {\color{blue} \textbf{46.1}} & {\color{blue} \textbf{24.9}} & {\color{blue} \textbf{46.4}} & {\color{blue} \textbf{54.8}} & 1.8 \\
				\midrule
				ResNet-50 &  \multirow{10}{*}{RetinaNet}  & 35.6 & 55.5 & 38.2 & 20.0 & 39.6 & 46.8 &  \\
				+RLA (Ours) &  & \textbf{37.9} & 57.0 & \textbf{40.8} & \textbf{22.0} & \textbf{41.7} & 49.2 & 2.3 \\
				+SE &  & 37.1 & 57.2 & 39.9 & 21.2 & 40.7 & 49.3 &  \\
				+ECA &  & 37.3 & 57.7 & 39.6 & 21.9 & 41.3 & 48.9 &  \\
				+ECA+RLA (Ours) &  & {\color{blue} \textbf{38.9}} & {\color{blue}\textbf{58.7}}  & {\color{blue} \textbf{41.7}} & {\color{blue} \textbf{23.9}} & {\color{blue} \textbf{42.7}} & {\color{blue} \textbf{49.7}} & 1.6 \\
				\cmidrule{1-1} \cmidrule{3-9}
				ResNet-101 &  & 37.7 & 57.5 & 40.4 & 21.1 & 42.2 & 49.5 &  \\
				+RLA (Ours) &  & \textbf{40.3} & {59.8} & \textbf{43.5} & {\textbf{24.2}} & \textbf{43.8} & \textbf{52.7} & 2.6 \\
				+SE &  & 38.7 & 59.1 & 41.6 & 22.1 & 43.1 & 50.9 &  \\
				+ECA &  & 39.1 & 59.9 & 41.8 & 22.8 & 43.4 & 50.6 &  \\
				+ECA+RLA (Ours) &  & {\color{blue} \textbf{41.5}} & {\color{blue}\textbf{61.6}}  & {\color{blue} \textbf{44.4}} & {\color{blue} \textbf{25.3}} & {\color{blue} \textbf{45.7}} & {\color{blue} \textbf{53.8}} & 2.4 \\
				\bottomrule
			\end{tabular}
		\end{small}
	\end{center}
	\vskip -0.2cm
\end{table}

\begin{table}[t]
	\caption{Object detection and instance segmentation results of different methods using Mask R-CNN on COCO val2017. $AP^{bb}$ and $AP^{m}$ denote AP of bounding box detection and instance segmentation.}
	\vskip -0.1in
	\label{tab:coco_mask-rcnn}
	\begin{center}
		\begin{small}
			\begin{tabular}{l|ccc|ccc|c|c}
				\toprule
				Methods & $AP^{bb}$  &  $AP_{50}^{bb}$  &  $AP_{75}^{bb}$  &  $AP^{m}$  &  $AP_{50}^{m}$  &  $AP_{75}^{m}$ & $\uparrow AP^{bb}$ & $\uparrow AP^{m}$ \\
				\midrule
				ResNet-50 & 37.2 & 58.9 & 40.3 & 34.1 & 55.5 & 36.2 &  &  \\
				+RLA (Ours) & \textbf{39.5} & 60.1 & \textbf{43.3} & \textbf{35.6} & 56.9 & \textbf{38.0} & 2.3 & 1.5 \\
				+SE & 38.7 & 60.9 & 42.1 & 35.4 & 57.4 & 37.8 &  &  \\
				+1 NL & 38.0 & 59.8 & 41.0 & 34.7 & 56.7 & 36.6 &  &  \\
				+GC Block & 39.4 & 61.6 & 42.4 & 35.7 & {{58.4}} & 37.6 &  &  \\
				+ECA & 39.0 & 61.3 & 42.1 & 35.6 & 57.1 & 37.7 &  &  \\
				+ECA+RLA (Ours) & {\color{blue} \textbf{40.6}} & {\color{blue} \textbf{61.8}} & {\color{blue} \textbf{44.0}} & {\color{blue} \textbf{36.5}} & {\color{blue}\textbf{58.4}} & {\color{blue} \textbf{38.8}} & 1.6 & 0.9 \\
				\midrule
				ResNet-101 & 39.4 & 60.9 & 43.3 & 35.9 & 57.7 & 38.4 &  &  \\
				+RLA (Ours) & \textbf{41.8} & 62.3 & \textbf{46.2} & 37.3 & 59.2 & \textbf{40.1} & 2.4 & 1.4 \\
				+SE & 40.7 & 62.5 & 44.3 & 36.8 & 59.3 & 39.2 &  &  \\
				+ECA & 41.3 & 63.1 & 44.8 & 37.4 & 59.9 & 39.8 &  &  \\
				+ECA+RLA (Ours) & {\color{blue} \textbf{42.9}} & {\color{blue} \textbf{63.6}} & {\color{blue} \textbf{46.9}} & {\color{blue} \textbf{38.1}} & {\color{blue} \textbf{60.5}} & {\color{blue} \textbf{40.8}} & 1.6 & 0.7 \\
				\bottomrule
			\end{tabular}
		\end{small}
	\end{center}
	\vskip -0.2cm
\end{table}

To show the transferability and the generalization ability, we experiment our RLA-Net on object detection task using Faster R-CNN \cite{fasterrcnn2015}, Mask R-CNN \cite{he2017mask} and RetinaNet \cite{lin2017focal} as detectors.
For Mask R-CNN, we also show instance segmentation results.
All detectors are implemented by the open source MMDetection toolkit \cite{mmdetection}.
We employ the same settings as in \cite{wang2020eca} to finetune our RLA-Nets on COCO train2017 set.
\begin{revised}Specifically, the shorter side of input images are resized to 800. 
We train all detectors within 12 epochs using SGD with weight decay of 1e-4, momentum of 0.9 and mini-batch size of 8. 
The learning rate is initialized to 0.01 and is decreased by a factor of 10 after 8 and 11 epochs, respectively, i.e., the 1x training schedule.\end{revised}
Different from the classification task, we do not pass the hidden states of our RLA module to the FPN.
Thus, the gains are solely from the more powerful representations learnt in the main CNNs.

We report the results on COCO val2017 set by standard COCO metrics of Average Precision (AP).
Tables \ref{tab:coco_detection} and \ref{tab:coco_mask-rcnn} report the performances of the object detectors. 
We can observe that our RLA-Net outperforms the original ResNet by 2.4\% and 2.5\% in terms of AP for the networks of 50 and 101 layers, respectively. 
More excitingly, our RLA-ResNet-50 outperforms ResNet-101 on these three detectors.
In particular, our RLA module achieves more gains in $AP_{75}$ and $AP_{L}$, indicating the high accuracy and effectiveness for large objects.
Remarkably, exploiting RLA on ECANets can surpass all other models with large margins.
We hypothesize that, with the help of RLA modules, positional information from low-level features is better preserved in the main CNN, leading to these notable improvements. 
In summary, the results in Tables \ref{tab:coco_detection} and \ref{tab:coco_mask-rcnn} demonstrate that our RLA-Net can well generalize to various tasks with extraordinary benefits on the object detection task.

\begin{revised}
	
\subsection{Ablation study}
We conduct an ablation study on ImageNet with ResNet-50 as the baseline network. 
To validate the effectiveness of the detailed inner structure, we ablate the important design elements of RLA module and compare the proposed structure with the following variants:

\quad~ (a) channels increased by 32;

\quad~ (b) RLA without parameter sharing;

\quad~ (c) RLA without the information flow from RLA to the main CNN (as in \cite{moniz2016convolutional});

\quad~ (d) RLA with ConvLSTM cell;

\quad~ (e) RLA with the post-activated hidden unit; and 

\quad (f-j) RLA with different connectivity between two paths, see Figure \ref{fig:ablation}.

The results are reported in Table \ref{tab:ablation-structure}.
We first demonstrate that the improvement in accuracy is not fully caused by the increment in network width by comparing to the variant with 32 additional channels in each layer. 
We then investigate the necessity of recurrence in our module by comparing it with its fully unshared counterpart.
Results show that the recurrent design can reduce parameter count and slightly increase accuracy at the same time.
Removing the information flow from RLA to the main CNN results in a decrease in model performances, validating that the information exchange between RLA and the main CNN is a must.
Furthermore, using ConvLSTM cells may be unnecessary since each stage only has 3 to 6 time steps in ResNet-50, and the post-activated recurrent unit is not very compatible with the pre-activated network.

We also compare different integration strategies to combine our RLA module with the main CNN. 
In addition to the proposed design, we consider five variants (v2-v6), as depicted in Figure \ref{fig:ablation}.
The performances of the variants are reported in Table \ref{tab:ablation-structure}. 
We observe that the variants v1, v3 and v5 perform similarly well, as they all integrate new information from the main CNNs before entering into the recurrent operation. 
Compared with variants v3 and v5, v1 obtains the learned residual as new information, including less redundancy.
The performance of our RLA modules is relatively robust to the integration strategies, as long as new information is added before the recurrent operation.

In summary, the proposed RLA module structure in Figure \ref{fig:RLA} achieves the smallest top-1 error and very competitive top-5 error with the minimum number of additional parameters.
This structure can strike a good balance between the increase of computational cost and performance gain.
We also conduct a more comprehensive ablation study on the CIFAR-10 dataset with ResNet-164 as the baseline network.
Due to limited space, we present details of the ablation studies and the investigated module structures in the supplementary material B.4.

\begin{figure}[t]
	\begin{center}
		\centerline{\includegraphics[width=\linewidth]{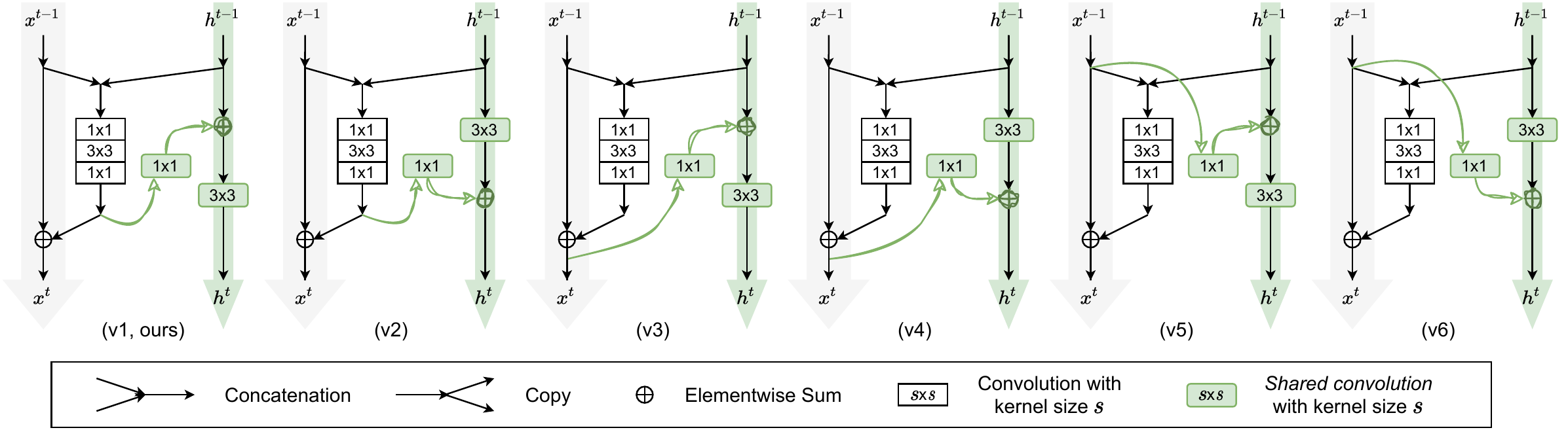}}
		\caption{RLA modules with different types of connections between two paths.}
		\label{fig:ablation}
	\end{center}
\end{figure}

\begin{table}[t]
	\caption{Classification errors on the ImageNet validation set using ResNet-50 as the baseline model.}
	\label{tab:ablation-structure}
	\centering
	\small
	\begin{tabular}{lcccc}
		\toprule
		Model  			&  Params (M)  &  FLOPs (G) & Top-1 err. & Top-5 err. \\
		\midrule
		ResNet-50 		& 24.37 & 3.83 & 24.70 & 7.80 \\
		(a) channel +32   & 25.55 & 4.24 & 23.54 & 6.72 \\
		(b) RLA-v1 ($k=32$, unshared) & 25.12 & 4.17 & 22.93 & 6.61 \\
		(c) RLA-v1 ($k=32$, no exchange) & 24.56 & 4.11 & 23.42 & 6.90 \\
		(d) RLA-v1 ($k=32$, ConvLSTM) & 24.92 & 5.00 & 22.92 & 6.53 \\
		(e) RLA-v1 ($k=32$, PostAct.) & 24.67 & 4.17 & 23.11 & 6.65 \\
		\midrule
		Proposed {RLA-v1 ($k=32$)}  & 24.67 & 4.17 &  {22.83}  & {6.58}  \\
		(f) RLA-v2 ($k=32$) 	    & 24.67 & 4.17 &  23.24  & 6.61  \\
		(g) RLA-v3 ($k=32$) 	    & 24.67 & 4.17 &  22.95  & 6.54   \\
		(h) RLA-v4 ($k=32$) 		& 24.67 & 4.17 &  23.36  & 6.72  \\
		(i) RLA-v5 ($k=32$)		& 24.67 & 4.17 &  23.30  & 6.60  \\
		(j) RLA-v6 ($k=32$)		& 24.67 & 4.17 &  23.50  & 6.82  \\
		\bottomrule
	\end{tabular}
\end{table}

\end{revised}

\section{Conclusion and future work} \label{sec:conclusion}
This paper proposes a recurrent layer aggregation module that is light-weighted and capable of providing the backbone CNN with sequentially aggregated information from previous layers.
Its potential has been well illustrated by the applications on mainstream CNNs with benchmark datasets. 
On the other hand, we mainly focused on controlled experiments to illustrate the effectiveness and compatibility of the RLA modules with settings adopted from existing work.
Our first future work is to consider hyperparameter tuning, advanced training tricks \cite{he2019bag, bello2021revisiting}, or architecture search \cite{tan2019efficientnet}.
Moreover, in terms of RLA design and application, it is also an interesting direction to apply it at the stage level on top of the layer level, which leads to deep RLA modules.

\begin{ack}
%
We thank the anonymous reviewers for their constructive feedback and Jinxu Zhao for technical support.
This project is supported by National Supercomputer Center in Guangzhou, China.
\end{ack}


%
%
%
%

\bibliography{RLA}

\begin{thebibliography}{63}
\providecommand{\natexlab}[1]{#1}
\providecommand{\url}[1]{\texttt{#1}}
\expandafter\ifx\csname urlstyle\endcsname\relax
  \providecommand{\doi}[1]{doi: #1}\else
  \providecommand{\doi}{doi: \begingroup \urlstyle{rm}\Url}\fi

\bibitem[Batsos and Mordohai(2018)]{batsos2018recresnet}
Konstantinos Batsos and Philippos Mordohai.
\newblock Recresnet: A recurrent residual cnn architecture for disparity map
  enhancement.
\newblock In \emph{2018 International Conference on 3D Vision (3DV)}, pages
  238--247. IEEE, 2018.

\bibitem[Bello et~al.(2021)Bello, Fedus, Du, Cubuk, Srinivas, Lin, Shlens, and
  Zoph]{bello2021revisiting}
Irwan Bello, William Fedus, Xianzhi Du, Ekin~D Cubuk, Aravind Srinivas,
  Tsung-Yi Lin, Jonathon Shlens, and Barret Zoph.
\newblock Revisiting resnets: Improved training and scaling strategies.
\newblock \emph{arXiv preprint arXiv:2103.07579}, 2021.

\bibitem[Chang et~al.(2017)Chang, Zhang, Han, Yu, Guo, Tan, Cui, Witbrock,
  Hasegawa-Johnson, and Huang]{chang2017dilated}
Shiyu Chang, Yang Zhang, Wei Han, Mo~Yu, Xiaoxiao Guo, Wei Tan, Xiaodong Cui,
  Michael Witbrock, Mark Hasegawa-Johnson, and Thomas~S Huang.
\newblock Dilated recurrent neural networks.
\newblock In \emph{Advances in Neural Information Processing Systems}, 2017.

\bibitem[Chen et~al.(2019)Chen, Wang, Pang, Cao, Xiong, Li, Sun, Feng, Liu, Xu,
  Zhang, Cheng, Zhu, Cheng, Zhao, Li, Lu, Zhu, Wu, Dai, Wang, Shi, Ouyang, Loy,
  and Lin]{mmdetection}
Kai Chen, Jiaqi Wang, Jiangmiao Pang, Yuhang Cao, Yu~Xiong, Xiaoxiao Li,
  Shuyang Sun, Wansen Feng, Ziwei Liu, Jiarui Xu, Zheng Zhang, Dazhi Cheng,
  Chenchen Zhu, Tianheng Cheng, Qijie Zhao, Buyu Li, Xin Lu, Rui Zhu, Yue Wu,
  Jifeng Dai, Jingdong Wang, Jianping Shi, Wanli Ouyang, Chen~Change Loy, and
  Dahua Lin.
\newblock {MMDetection}: Open mmlab detection toolbox and benchmark.
\newblock \emph{arXiv preprint arXiv:1906.07155}, 2019.

\bibitem[Chen et~al.(2017{\natexlab{a}})Chen, Zhang, Xiao, Nie, Shao, Liu, and
  Chua]{chen2017sca}
Long Chen, Hanwang Zhang, Jun Xiao, Liqiang Nie, Jian Shao, Wei Liu, and
  Tat-Seng Chua.
\newblock Sca-cnn: Spatial and channel-wise attention in convolutional networks
  for image captioning.
\newblock In \emph{Proceedings of the IEEE Conference on Computer Vision and
  Pattern Recognition}, pages 5659--5667, 2017{\natexlab{a}}.

\bibitem[Chen et~al.(2018)Chen, Liu, Gao, and Han]{chen2018mobilefacenets}
Sheng Chen, Yang Liu, Xiang Gao, and Zhen Han.
\newblock Mobilefacenets: Efficient cnns for accurate real-time face
  verification on mobile devices.
\newblock In \emph{Chinese Conference on Biometric Recognition}, pages
  428--438. Springer, 2018.

\bibitem[Chen et~al.(2017{\natexlab{b}})Chen, Li, Xiao, Jin, Yan, and
  Feng]{chen2017dual}
Yunpeng Chen, Jianan Li, Huaxin Xiao, Xiaojie Jin, Shuicheng Yan, and Jiashi
  Feng.
\newblock Dual path networks.
\newblock In \emph{Advances in Neural Information Processing Systems},
  2017{\natexlab{b}}.

\bibitem[Cho et~al.(2014)Cho, Van~Merri{\"e}nboer, Gulcehre, Bahdanau,
  Bougares, Schwenk, and Bengio]{cho2014learning}
Kyunghyun Cho, Bart Van~Merri{\"e}nboer, Caglar Gulcehre, Dzmitry Bahdanau,
  Fethi Bougares, Holger Schwenk, and Yoshua Bengio.
\newblock Learning phrase representations using rnn encoder-decoder for
  statistical machine translation.
\newblock In \emph{EMNLP}, 2014.

\bibitem[Choi et~al.(2020)Choi, Kim, Han, Xu, and Lee]{choi2020channel}
Myungsub Choi, Heewon Kim, Bohyung Han, Ning Xu, and Kyoung~Mu Lee.
\newblock Channel attention is all you need for video frame interpolation.
\newblock In \emph{Proceedings of the AAAI Conference on Artificial
  Intelligence}, volume~34, pages 10663--10671, 2020.

\bibitem[Chollet(2017)]{chollet2017xception}
Fran{\c{c}}ois Chollet.
\newblock Xception: Deep learning with depthwise separable convolutions.
\newblock In \emph{Proceedings of the IEEE Conference on Computer Vision and
  Pattern Recognition}, pages 1251--1258, 2017.

\bibitem[Connor et~al.(1994)Connor, Martin, and Atlas]{connor1994recurrent}
Jerome~T Connor, R~Douglas Martin, and Les~E Atlas.
\newblock Recurrent neural networks and robust time series prediction.
\newblock \emph{IEEE transactions on neural networks}, 5\penalty0 (2):\penalty0
  240--254, 1994.

\bibitem[Deng et~al.(2009)Deng, Dong, Socher, Li, Li, and
  Fei-Fei]{deng2009imagenet}
Jia Deng, Wei Dong, Richard Socher, Li-Jia Li, Kai Li, and Li~Fei-Fei.
\newblock Imagenet: A large-scale hierarchical image database.
\newblock In \emph{Proceedings of the IEEE Conference on Computer Vision and
  Pattern Recognition}, pages 248--255. Ieee, 2009.

\bibitem[DiPietro et~al.(2017)DiPietro, Rupprecht, Navab, and
  Hager]{dipietro2017analyzing}
Robert DiPietro, Christian Rupprecht, Nassir Navab, and Gregory~D Hager.
\newblock Analyzing and exploiting narx recurrent neural networks for long-term
  dependencies.
\newblock \emph{arXiv preprint arXiv:1702.07805}, 2017.

\bibitem[Donahue et~al.(2015)Donahue, Anne~Hendricks, Guadarrama, Rohrbach,
  Venugopalan, Saenko, and Darrell]{donahue2015long}
Jeffrey Donahue, Lisa Anne~Hendricks, Sergio Guadarrama, Marcus Rohrbach,
  Subhashini Venugopalan, Kate Saenko, and Trevor Darrell.
\newblock Long-term recurrent convolutional networks for visual recognition and
  description.
\newblock In \emph{Proceedings of the IEEE Conference on Computer Vision and
  Pattern Recognition}, pages 2625--2634, 2015.

\bibitem[El~Hihi and Bengio(1996)]{el1996hierarchical}
Salah El~Hihi and Yoshua Bengio.
\newblock Hierarchical recurrent neural networks for long-term dependencies.
\newblock In \emph{Advances in Neural Information Processing Systems}, pages
  493--499, 1996.

\bibitem[Fu et~al.(2019)Fu, Liu, Tian, Li, Bao, Fang, and Lu]{fu2019dual}
Jun Fu, Jing Liu, Haijie Tian, Yong Li, Yongjun Bao, Zhiwei Fang, and Hanqing
  Lu.
\newblock Dual attention network for scene segmentation.
\newblock In \emph{Proceedings of the IEEE Conference on Computer Vision and
  Pattern Recognition}, pages 3146--3154, 2019.

\bibitem[Greff et~al.(2017)Greff, Srivastava, and
  Schmidhuber]{greff2017highway}
Klaus Greff, Rupesh~K Srivastava, and J{\"u}rgen Schmidhuber.
\newblock Highway and residual networks learn unrolled iterative estimation.
\newblock In \emph{International Conference on Learning Representations}, 2017.

\bibitem[He et~al.(2015{\natexlab{a}})He, Zhang, Ren, and Sun]{he2015delving}
Kaiming He, Xiangyu Zhang, Shaoqing Ren, and Jian Sun.
\newblock Delving deep into rectifiers: Surpassing human-level performance on
  imagenet classification.
\newblock In \emph{Proceedings of the IEEE International Conference on Computer
  Vision}, pages 1026--1034, 2015{\natexlab{a}}.

\bibitem[He et~al.(2015{\natexlab{b}})He, Zhang, Ren, and Sun]{resnetgit}
Kaiming He, Xiangyu Zhang, Shaoqing Ren, and Jian Sun.
\newblock Deep residual learning for image recognition.
\newblock \url{https://github.com/Kaiminghe/deep-residual-networks},
  2015{\natexlab{b}}.

\bibitem[He et~al.(2016{\natexlab{a}})He, Zhang, Ren, and Sun]{he2016deep}
Kaiming He, Xiangyu Zhang, Shaoqing Ren, and Jian Sun.
\newblock Deep residual learning for image recognition.
\newblock In \emph{Proceedings of the IEEE Conference on Computer Vision and
  Pattern Recognition}, pages 770--778, 2016{\natexlab{a}}.

\bibitem[He et~al.(2016{\natexlab{b}})He, Zhang, Ren, and Sun]{he2016identity}
Kaiming He, Xiangyu Zhang, Shaoqing Ren, and Jian Sun.
\newblock Identity mappings in deep residual networks.
\newblock In \emph{European Conference on Computer Vision}, pages 630--645.
  Springer, 2016{\natexlab{b}}.

\bibitem[He et~al.(2017)He, Gkioxari, Doll{\'a}r, and Girshick]{he2017mask}
Kaiming He, Georgia Gkioxari, Piotr Doll{\'a}r, and Ross Girshick.
\newblock Mask {R-CNN}.
\newblock In \emph{Proceedings of the IEEE international conference on computer
  vision}, pages 2961--2969, 2017.

\bibitem[He et~al.(2019)He, Zhang, Zhang, Zhang, Xie, and Li]{he2019bag}
Tong He, Zhi Zhang, Hang Zhang, Zhongyue Zhang, Junyuan Xie, and Mu~Li.
\newblock Bag of tricks for image classification with convolutional neural
  networks.
\newblock In \emph{Proceedings of the IEEE/CVF Conference on Computer Vision
  and Pattern Recognition}, pages 558--567, 2019.

\bibitem[Hochreiter and Schmidhuber(1997)]{hochreiter1997long}
Sepp Hochreiter and J{\"u}rgen Schmidhuber.
\newblock Long short-term memory.
\newblock \emph{Neural Computation}, 9\penalty0 (8):\penalty0 1735--1780, 1997.

\bibitem[Howard et~al.(2019)Howard, Sandler, Chu, Chen, Chen, Tan, Wang, Zhu,
  Pang, Vasudevan, et~al.]{howard2019searching}
Andrew Howard, Mark Sandler, Grace Chu, Liang-Chieh Chen, Bo~Chen, Mingxing
  Tan, Weijun Wang, Yukun Zhu, Ruoming Pang, Vijay Vasudevan, et~al.
\newblock Searching for mobilenetv3.
\newblock In \emph{Proceedings of the IEEE/CVF International Conference on
  Computer Vision}, pages 1314--1324, 2019.

\bibitem[Hu et~al.(2018)Hu, Shen, and Sun]{hu2018squeeze}
Jie Hu, Li~Shen, and Gang Sun.
\newblock Squeeze-and-excitation networks.
\newblock In \emph{Proceedings of the IEEE Conference on Computer Vision and
  Pattern Recognition}, pages 7132--7141, 2018.

\bibitem[Huang et~al.(2016)Huang, Sun, Liu, Sedra, and
  Weinberger]{huang2016deep}
Gao Huang, Yu~Sun, Zhuang Liu, Daniel Sedra, and Kilian~Q Weinberger.
\newblock Deep networks with stochastic depth.
\newblock In \emph{European Conference on Computer Vision}, pages 646--661.
  Springer, 2016.

\bibitem[Huang et~al.(2017)Huang, Liu, Van Der~Maaten, and
  Weinberger]{huang2017densely}
Gao Huang, Zhuang Liu, Laurens Van Der~Maaten, and Kilian~Q Weinberger.
\newblock Densely connected convolutional networks.
\newblock In \emph{Proceedings of the IEEE Conference on Computer Vision and
  Pattern Recognition}, pages 4700--4708, 2017.

\bibitem[Huang et~al.(2019)Huang, Liu, Pleiss, Van Der~Maaten, and
  Weinberger]{huang2019convolutional}
Gao Huang, Zhuang Liu, Geoff Pleiss, Laurens Van Der~Maaten, and Kilian
  Weinberger.
\newblock Convolutional networks with dense connectivity.
\newblock \emph{IEEE Transactions on Pattern Analysis and Machine
  Intelligence}, 2019.

\bibitem[Ioffe and Szegedy(2015)]{ioffe2015batch}
Sergey Ioffe and Christian Szegedy.
\newblock Batch normalization: Accelerating deep network training by reducing
  internal covariate shift.
\newblock In \emph{International Conference on Machine Learning}, pages
  448--456. PMLR, 2015.

\bibitem[Krizhevsky et~al.(2012)Krizhevsky, Sutskever, and
  Hinton]{krizhevsky2012imagenet}
Alex Krizhevsky, Ilya Sutskever, and Geoffrey~E Hinton.
\newblock Imagenet classification with deep convolutional neural networks.
\newblock \emph{Advances in Neural Information Processing Systems},
  25:\penalty0 1097--1105, 2012.

\bibitem[LeCun et~al.(2015)LeCun, Bengio, and Hinton]{lecun2015deep}
Yann LeCun, Yoshua Bengio, and Geoffrey Hinton.
\newblock Deep learning.
\newblock \emph{nature}, 521\penalty0 (7553):\penalty0 436--444, 2015.

\bibitem[Li et~al.(2018)Li, Xiong, An, and Wang]{li2018pyramid}
Hanchao Li, Pengfei Xiong, Jie An, and Lingxue Wang.
\newblock Pyramid attention network for semantic segmentation.
\newblock In \emph{British Machine Vision Conference}, 2018.

\bibitem[Li et~al.(2019)Li, Song, and Wu]{li2019aognets}
Xilai Li, Xi~Song, and Tianfu Wu.
\newblock Aognets: Compositional grammatical architectures for deep learning.
\newblock In \emph{Proceedings of the IEEE/CVF Conference on Computer Vision
  and Pattern Recognition}, pages 6220--6230, 2019.

\bibitem[Liang and Hu(2015)]{liang2015recurrent}
Ming Liang and Xiaolin Hu.
\newblock Recurrent convolutional neural network for object recognition.
\newblock In \emph{Proceedings of the IEEE Conference on Computer Vision and
  Pattern Recognition}, pages 3367--3375, 2015.

\bibitem[Liao and Poggio(2016)]{liao2016bridging}
Qianli Liao and Tomaso Poggio.
\newblock Bridging the gaps between residual learning, recurrent neural
  networks and visual cortex.
\newblock \emph{arXiv preprint arXiv:1604.03640}, 2016.

\bibitem[Lim and Lee(2017)]{lim2017deep}
Bee Lim and Kyoung~Mu Lee.
\newblock Deep recurrent resnet for video super-resolution.
\newblock In \emph{2017 Asia-Pacific Signal and Information Processing
  Association Annual Summit and Conference (APSIPA ASC)}, pages 1452--1455.
  IEEE, 2017.

\bibitem[Lin et~al.(2017{\natexlab{a}})Lin, Doll{\'a}r, Girshick, He,
  Hariharan, and Belongie]{lin2017fpn}
Tsung-Yi Lin, Piotr Doll{\'a}r, Ross Girshick, Kaiming He, Bharath Hariharan,
  and Serge Belongie.
\newblock Feature pyramid networks for object detection.
\newblock In \emph{Proceedings of the IEEE conference on computer vision and
  pattern recognition}, pages 2117--2125, 2017{\natexlab{a}}.

\bibitem[Lin et~al.(2017{\natexlab{b}})Lin, Goyal, Girshick, He, and
  Doll{\'a}r]{lin2017focal}
Tsung-Yi Lin, Priya Goyal, Ross Girshick, Kaiming He, and Piotr Doll{\'a}r.
\newblock Focal loss for dense object detection.
\newblock In \emph{Proceedings of the IEEE international conference on computer
  vision}, pages 2980--2988, 2017{\natexlab{b}}.

\bibitem[Lin et~al.(1996)Lin, Horne, Tino, and Giles]{lin1996learning}
Tsungnan Lin, Bill~G Horne, Peter Tino, and C~Lee Giles.
\newblock Learning long-term dependencies in narx recurrent neural networks.
\newblock \emph{IEEE Transactions on Neural Networks}, 7\penalty0 (6):\penalty0
  1329--1338, 1996.

\bibitem[Moniz and Pal(2016)]{moniz2016convolutional}
Joel Moniz and Christopher Pal.
\newblock Convolutional residual memory networks.
\newblock \emph{arXiv preprint arXiv:1606.05262}, 2016.

\bibitem[Ren et~al.(2015)Ren, He, Girshick, and Sun]{fasterrcnn2015}
Shaoqing Ren, Kaiming He, Ross Girshick, and Jian Sun.
\newblock Faster {R-CNN}: Towards real-time object detection with region
  proposal networks.
\newblock In \emph{Proceedings of the 28th International Conference on Neural
  Information Processing Systems - Volume 1}, NIPS'15, page 91–99, Cambridge,
  MA, USA, 2015. MIT Press.

\bibitem[Sandler et~al.(2018)Sandler, Howard, Zhu, Zhmoginov, and
  Chen]{sandler2018mobilenetv2}
Mark Sandler, Andrew Howard, Menglong Zhu, Andrey Zhmoginov, and Liang-Chieh
  Chen.
\newblock Mobilenetv2: Inverted residuals and linear bottlenecks.
\newblock In \emph{Proceedings of the IEEE Conference on Computer Vision and
  Pattern Recognition}, pages 4510--4520, 2018.

\bibitem[Shi et~al.(2016)Shi, Bai, and Yao]{shi2016end}
Baoguang Shi, Xiang Bai, and Cong Yao.
\newblock An end-to-end trainable neural network for image-based sequence
  recognition and its application to scene text recognition.
\newblock \emph{IEEE Transactions on Pattern Analysis and Machine
  Intelligence}, 39\penalty0 (11):\penalty0 2298--2304, 2016.

\bibitem[Shi et~al.(2015)Shi, Chen, Wang, Yeung, Wong, and
  Woo]{shi2015convolutional}
Xingjian Shi, Zhourong Chen, Hao Wang, Dit-Yan Yeung, Wai-Kin Wong, and
  Wang-chun Woo.
\newblock Convolutional lstm network: A machine learning approach for
  precipitation nowcasting.
\newblock In \emph{Advances in Neural Information Processing Systems}, 2015.

\bibitem[Szegedy et~al.(2015)Szegedy, Liu, Jia, Sermanet, Reed, Anguelov,
  Erhan, Vanhoucke, and Rabinovich]{szegedy2015going}
Christian Szegedy, Wei Liu, Yangqing Jia, Pierre Sermanet, Scott Reed, Dragomir
  Anguelov, Dumitru Erhan, Vincent Vanhoucke, and Andrew Rabinovich.
\newblock Going deeper with convolutions.
\newblock In \emph{Proceedings of the IEEE Conference on Computer Vision and
  Pattern Recognition}, pages 1--9, 2015.

\bibitem[Szegedy et~al.(2016)Szegedy, Vanhoucke, Ioffe, Shlens, and
  Wojna]{szegedy2016rethinking}
Christian Szegedy, Vincent Vanhoucke, Sergey Ioffe, Jon Shlens, and Zbigniew
  Wojna.
\newblock Rethinking the inception architecture for computer vision.
\newblock In \emph{Proceedings of the IEEE Conference on Computer Vision and
  Pattern Recognition}, pages 2818--2826, 2016.

\bibitem[Szegedy et~al.(2017)Szegedy, Ioffe, Vanhoucke, and
  Alemi]{szegedy2017inception}
Christian Szegedy, Sergey Ioffe, Vincent Vanhoucke, and Alexander Alemi.
\newblock Inception-v4, inception-resnet and the impact of residual connections
  on learning.
\newblock In \emph{Proceedings of the AAAI Conference on Artificial
  Intelligence}, volume~31, 2017.

\bibitem[Tan and Le(2019)]{tan2019efficientnet}
Mingxing Tan and Quoc Le.
\newblock Efficientnet: Rethinking model scaling for convolutional neural
  networks.
\newblock In \emph{International Conference on Machine Learning}, pages
  6105--6114. PMLR, 2019.

\bibitem[Vaswani et~al.(2017)Vaswani, Shazeer, Parmar, Uszkoreit, Jones, Gomez,
  Kaiser, and Polosukhin]{vaswani2017attention}
Ashish Vaswani, Noam Shazeer, Niki Parmar, Jakob Uszkoreit, Llion Jones,
  Aidan~N Gomez, {\L}ukasz Kaiser, and Illia Polosukhin.
\newblock Attention is all you need.
\newblock In \emph{Advances in Neural Information Processing Systems}, pages
  5998--6008, 2017.

\bibitem[Vinyals et~al.(2015)Vinyals, Toshev, Bengio, and
  Erhan]{vinyals2015show}
Oriol Vinyals, Alexander Toshev, Samy Bengio, and Dumitru Erhan.
\newblock Show and tell: A neural image caption generator.
\newblock In \emph{Proceedings of the IEEE Conference on Computer Vision and
  Pattern Recognition}, pages 3156--3164, 2015.

\bibitem[Wang et~al.(2017)Wang, Jiang, Qian, Yang, Li, Zhang, Wang, and
  Tang]{wang2017residual}
Fei Wang, Mengqing Jiang, Chen Qian, Shuo Yang, Cheng Li, Honggang Zhang,
  Xiaogang Wang, and Xiaoou Tang.
\newblock Residual attention network for image classification.
\newblock In \emph{Proceedings of the IEEE Conference on Computer Vision and
  Pattern Recognition}, pages 3156--3164, 2017.

\bibitem[Wang et~al.(2016)Wang, Yang, Mao, Huang, Huang, and Xu]{wang2016cnn}
Jiang Wang, Yi~Yang, Junhua Mao, Zhiheng Huang, Chang Huang, and Wei Xu.
\newblock Cnn-rnn: A unified framework for multi-label image classification.
\newblock In \emph{Proceedings of the IEEE Conference on Computer Vision and
  Pattern Recognition}, pages 2285--2294, 2016.

\bibitem[Wang et~al.(2020)Wang, Wu, Zhu, Li, Zuo, and Hu]{wang2020eca}
Qilong Wang, Banggu Wu, Pengfei Zhu, Peihua Li, Wangmeng Zuo, and Qinghua Hu.
\newblock Eca-net: Efficient channel attention for deep convolutional neural
  networks.
\newblock In \emph{Proceedings of the IEEE Conference on Computer Vision and
  Pattern Recognition}, 2020.

\bibitem[Wightman(2019)]{rw2019timm}
Ross Wightman.
\newblock Pytorch image models.
\newblock \url{https://github.com/rwightman/pytorch-image-models}, 2019.

\bibitem[Woo et~al.(2018)Woo, Park, Lee, and Kweon]{woo2018cbam}
Sanghyun Woo, Jongchan Park, Joon-Young Lee, and In~So Kweon.
\newblock Cbam: Convolutional block attention module.
\newblock In \emph{Proceedings of the European conference on computer vision
  (ECCV)}, pages 3--19, 2018.

\bibitem[Xie et~al.(2017)Xie, Girshick, Doll{\'a}r, Tu, and
  He]{xie2017aggregated}
Saining Xie, Ross Girshick, Piotr Doll{\'a}r, Zhuowen Tu, and Kaiming He.
\newblock Aggregated residual transformations for deep neural networks.
\newblock In \emph{Proceedings of the IEEE Conference on Computer Vision and
  Pattern Recognition}, pages 1492--1500, 2017.

\bibitem[Xu et~al.(2015)Xu, Ba, Kiros, Cho, Courville, Salakhudinov, Zemel, and
  Bengio]{xu2015show}
Kelvin Xu, Jimmy Ba, Ryan Kiros, Kyunghyun Cho, Aaron Courville, Ruslan
  Salakhudinov, Rich Zemel, and Yoshua Bengio.
\newblock Show, attend and tell: Neural image caption generation with visual
  attention.
\newblock In \emph{International Conference on Machine Learning}, pages
  2048--2057. PMLR, 2015.

\bibitem[Yang et~al.(2020)Yang, An, Zhu, Hu, Zhang, Xu, Li, and
  Xu]{yang2020gated}
Chuanguang Yang, Zhulin An, Hui Zhu, Xiaolong Hu, Kun Zhang, Kaiqiang Xu, Chao
  Li, and Yongjun Xu.
\newblock Gated convolutional networks with hybrid connectivity for image
  classification.
\newblock In \emph{Proceedings of the AAAI Conference on Artificial
  Intelligence}, volume~34, pages 12581--12588, 2020.

\bibitem[Yu et~al.(2018)Yu, Wang, Shelhamer, and Darrell]{yu2018deep}
Fisher Yu, Dequan Wang, Evan Shelhamer, and Trevor Darrell.
\newblock Deep layer aggregation.
\newblock In \emph{Proceedings of the IEEE Conference on Computer Vision and
  Pattern Recognition}, pages 2403--2412, 2018.

\bibitem[Zhang et~al.(2016)Zhang, Wu, Che, Lin, Memisevic, Salakhutdinov, and
  Bengio]{zhang2016architectural}
Saizheng Zhang, Yuhuai Wu, Tong Che, Zhouhan Lin, Roland Memisevic, Ruslan
  Salakhutdinov, and Yoshua Bengio.
\newblock Architectural complexity measures of recurrent neural networks.
\newblock In \emph{Advances in Neural Information Processing Systems}, 2016.

\bibitem[Zhao et~al.(2020)Zhao, Huang, Lv, Duan, Qin, Li, and
  Tian]{zhao2020rnn}
Jingyu Zhao, Feiqing Huang, Jia Lv, Yanjie Duan, Zhen Qin, Guodong Li, and
  Guangjian Tian.
\newblock Do rnn and lstm have long memory?
\newblock In \emph{International Conference on Machine Learning}, pages
  11365--11375. PMLR, 2020.

\bibitem[Zhu et~al.(2016)Zhu, Groth, Bernstein, and Fei-Fei]{zhu2016visual7w}
Yuke Zhu, Oliver Groth, Michael Bernstein, and Li~Fei-Fei.
\newblock Visual7w: Grounded question answering in images.
\newblock In \emph{Proceedings of the IEEE Conference on Computer Vision and
  Pattern Recognition}, pages 4995--5004, 2016.

\end{thebibliography}
\bibliographystyle{plainnat}


%
%

\newpage
\appendix

\section*{Appendix}
%

\section{Recurrent Layer Aggregation Modules} \label{sec:supp-RLA}

\subsection{Layer Aggregation}
This subsection provides more explanations of the layer aggregation mechanism.
In the main paper, Eq. (1) gives the general definition
\begin{equation*}
A^t  = g^t(x^{t-1},x^{t-2},\ldots,x^0) \hspace{5mm}\text{and}\hspace{5mm}
x^t  = f^t (A^{t-1}, x^{t-1}), 
\end{equation*}
and Eq. (2) considers the simplified additive form
\begin{equation*}
A^t  =  \tsum_{l=0}^{t-1} g_{l}^t (x^l),
\end{equation*}
which is often satisfied if we impose a linearity assumption and consider the corresponding network without nonlinear activation functions.
In the following, we revisit the examples in Section 3.1.

\paragraph{Example 1. Sequential CNNs without skip connections.}
Such CNNs do not fulfill Eq. (1) or (2) because the information in $x^l $ with $ l < t-1 $ is only available to $ x^t$ through $x^{t-1}$.
More rigorously, we may write the features at the $t$-th layer as 
\begin{align*}
x^t & = \conv_t (x^{t-1}) , \\
x^t & = \conv_t (\conv_{t-1} (x^{t-2})) , \\
& \vdots \\
x^t & = \conv_t (\conv_{t-1} ( \cdots \conv_{1} (x^{0}))),
\end{align*}
which do not have the form of Eq. (1) or (2). 
Here, $ \conv_l(\cdot) $ for $1\leq l \leq t, $ denotes convolution operations at the $l$th layer.

\paragraph{Example 2. Hierarchical Layer Aggregation (HLA) \cite{yu2018deep}.} 
HLA satisfies Eq. (1) since its base aggregation operation $N$ (see Eq. 3 in \cite{yu2018deep}) is given by
\begin{equation*}
N(x^1, ..., x^n) = \sigma(\text{BatchNorm}(\sum_i W_i x^i +b )).
\end{equation*}
Furthermore, if we ignore nonlinearity $\sigma$ and batch normalization and rewrite weights multiplication as a convolution function $ \conv_i(\cdot) $, we have 
\begin{equation*}
N(x^1, ..., x^n) = \sum_i \conv_i(x^i) ,
\end{equation*}
which takes the form of Eq. (2).

\paragraph{Example 3. DenseNets.} 
According to \cite{chen2017dual}, Eq. (4)
\begin{equation*}
\convo^t (\text{Concat}(x^0, x^1, ..., x^{t-1})) = \tsum_{l=0}^{t-1} \convo_{l}^t (x^{l})
\end{equation*}
holds when dense connectivity is followed by a 1x1 convolution.
In Section 3.1, we argued that DenseNets implement channel attention across layers in the sense of \cite{vaswani2017attention}.
For a DenseNet with growth rate parameter $k$, the corresponding multi-head attention mechanism in Eq. (5) has $m=4k$ heads, where each head is a weighted average of all channels in previous layers.
This interpretation is consistent with the design of attention mechanisms, while a noticeable difference is that the similarity measures are learned in the convolution $ \convo^t $.


Moreover, changing the $ \convo^t $ in DenseNets to a convolution with kernel size other than 1x1	will render the weighted average formulation impossible and invalidate its attention interpretation.
In the meanwhile, we claimed that such modifications preserve the layer aggregation mechanism.
Consider a modified DenseNet layer
\begin{equation*}
x^t = \convt^t \left[\conv^t (\text{Concat}(x^0, x^1, ..., x^{t-1})) \right] ,
\end{equation*}
which can still be rewritten as
\begin{equation*}
A^t = \tsum_{l=0}^{t-1} \conv_{l}^{t+1} (x^l) \hspace{5mm}\text{and}\hspace{5mm}
x^t  = \convt^t \left[ A^{t-1} + \conv_{t-1}^t (x^{t-1}) \right].
\end{equation*}
Thus, the modified DenseNet layer still implements the layer aggregation mechanism.

\subsection{Sequential structure of layers in deep networks} \label{sec:time-series}


\paragraph{Time Series Background}
\begin{revised}
	In the literature of time series analysis, AR and ARMA models are most commonly used to fit and to predict time series data. 
	For example, given the historical traffic volume $ z_1, z_2, ..., z_{t-1} $ on a specific road, the government wants to predict future traffic $ z_t, z_{t+1}, ... $. 
	Generally speaking, each observation $ z_t $ can depend on all previous ones in a nonlinear manner and the relationship may vary as $t$ differs. 
	AR and ARMA models are linear and assume a formulation where coefficients only depend on the \textit{time lag} between the target and the past observation, shared across different target timestamp $t$. 
	Thus, they are simple, parsimonious and were proved to be useful in many applications.
	
	Time series is a sequence of random variables and the white noise sequence plays a necessary role in time series models as the source of randomness. 
	In the main text, we refer to the white noise term as the additive error because it is an addend in the model equation.
\end{revised}

Formally, a $p$th order autoregressive model, a.k.a., an AR($p$) model, $ \{Z_t\} $
satisfies
\begin{equation*}
Z_t = \theta_0 + \phi_1Z_{t-1} + \phi_2Z_{t-2} + \cdots + \phi_pZ_{t-p} + a_t,
\end{equation*}
where $ p\geq 0 $ is an integer, $ \phi$'s are real parameters and $ \{a_t\} $ is a white noise sequence.
An autoregressive-moving-average model of orders $p$ and $q$, a.k.a., an ARMA$ (p, q) $ model, satisfies 
\begin{equation*}
Z_t = \theta_0 + \phi_1Z_{t-1} + \phi_2Z_{t-2} + \cdots + \phi_pZ_{t-p} + a_t - \theta_1a_{t-1} - \theta_2 a_{t-1} - \cdots - \theta_q a_{t-q},
\end{equation*}
where $ p, q\geq 0 $ are integers, $ \theta$'s are real parameters for the MA part.
\textit{Invertibility} is a property of time series models to characterize whether the information sequence $ \{a_t\} $ can be recovered from past observations, and it is always required in time series applications.
If an ARMA model is invertible, then it has an \textit{AR representation}
\begin{equation*}
Z_t = \pi_1 Z_{t-1} + \pi_2 Z_{t-2} + \cdots + a_t ,
\end{equation*}
which takes the form of an AR($ \infty$) model, where the infinitely many $\pi$'s can be fully determined by $ p+q $ parameters, i.e., $\phi$'s and $\theta$'s.
In the main text, we claim that ARMA model simplifies AR and give an example using ARMA$(1, 1)$; see Eq. (7).
We are referring to the fact that any invertible ARMA model has an AR representation, whose parameters can be fully determined by the (much fewer) parameters of the ARMA model.

\paragraph{Implementation of two simplifications}

\begin{figure}[ht]
	\centering
	\includegraphics{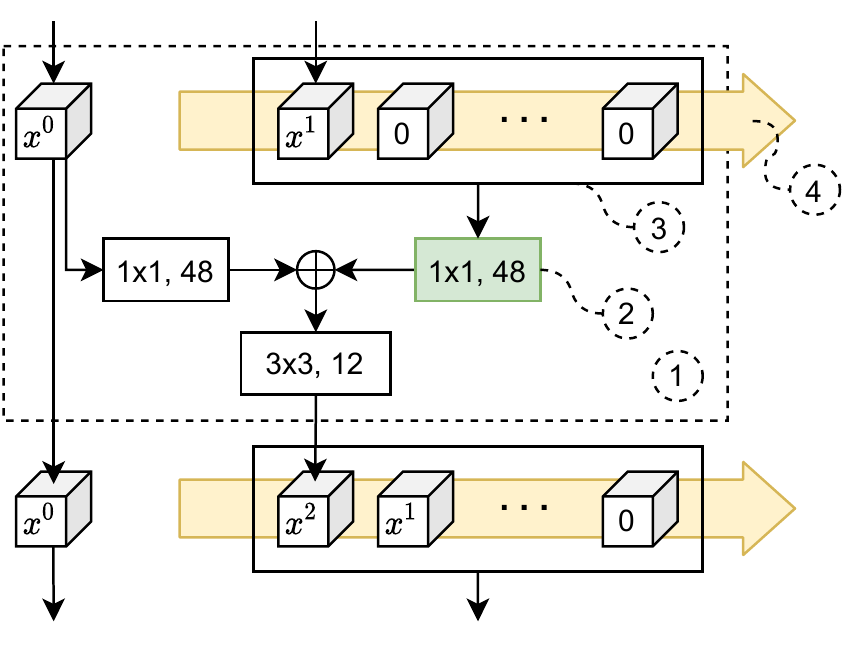}
	\caption{Visualization of a layer in the {Shared-Lag} modification of DenseNet in Eq. (9).}	
	\label{fig:shared1x1}
\end{figure}

Figure \ref{fig:shared1x1} visualizes the structure of a modified DenseNet layer following Eq. (9), where
cubes represent feature tensors, 
rectangles are convolutions annotated with kernel size and output channels, 
and $ \oplus $ denotes elementwise addition.
Numbers inside dashed circles correspond to the following notes: 
\begin{enumerate}
	\item Operations inside the large dashed rectangle represents the $2$nd layer in a dense block.
	\item This 1x1 convolution is shared among layers within a stage.
	\item Dense connections are implemented by this storage area of past layers' output, and $0$ is used as a placeholder.
	\item This arrow represents a virtual "conveyor belt" indicating that the feature maps are always ordered from the most recent layer to the most distant layer.
\end{enumerate}
$ \convt^t $ and $ \convo_{0}^t(\cdot) $ in Eq. (9) are represented by the unshared 3x3 and 1x1 convolutions (white rectangles), and a combination of $ \convo_1, \convo_2, ..., \convo_{15} $ is represented by the shared 1x1 convolution (colored rectangle with note 2).
Feature maps from previous layers are ordered such that Eq. (9) holds.
The Shared-Ordinal variant of DenseNet can be similarly implemented by adjusting the ordering (see note 3 above) and disabling the "conveyor belt" (see note 4 above). 

\paragraph{Weights of the shared 1x1 convolutions}
Figure 2 in the main paper shows the $ L_1 $ norm of the weights of the shared 1x1 convolutions. 
And we observe the following main difference between the two modes of weight sharing:
\begin{itemize}
	\item When weights are shared based on lag, \textit{the most recent layer} corresponds to the largest weight, and the $ L_1$ norm decays as lag increases. 
	\item When weights are shared according to the ordinal number of the layers, \textit{layers closer to the input} correspond to larger weights, and the first few layers have similar weights.
\end{itemize}
We also claimed that the quick decaying pattern in the plot for Shared-Lag could be exponential decay, and we provide numerical support below.
We fit exponential functions to the curves in Figure 2 and use the $ R $-squared metric to compare goodness-of-fit. 
$ R^2 \in (0, 1) $ is a metric that can be interpreted as the proportion of variance explained by a model, i.e., a fitted curve. 
$ R^2$ values for the three curves in the Share-Lag subplot are $ 0.95, 0.86, 0.90 $, while that for Shared-Ordinal are $ 0.77, 0.91, 0.87 $, implying poorer fit.

\subsection{Recurrent Layer Aggregation}
This subsection shows that the RLA module in Eq. (11) implements layer aggregation as in Eq. (1). 
And under a linearity assumption, Eq. (11) can also be shown to have the additive form in Eq. (2).

Recall the formulation of RLA mechanism in Eq. (11), 
\begin{equation*}
h^t = g^t (h^{t-1}, x^{t-1})\hspace{5mm}\text{and}\hspace{5mm}
x^t=f^t (h^{t-1}, x^{t-1}).
\end{equation*}
Recursively substitute $ h^s = g^s (h^{s-1}, x^{s-1}) $ for $ s = t-1, t-2, ... $ into the first equation, we have
\begin{align*}
h^t & = g^t (h^{t-1}, x^{t-1}) \\
& = g^t (g^{t-1} (h^{t-2}, x^{t-2}), x^{t-1}) \\
& = \cdots,
\end{align*}
which is a function of $ x^{t-1}, x^{t-2}, ..., x^0 $ and a constant $ h^0 $.
Thus, Eq. (1) is satisfied.

If we further assume that there exist functions $ g_1^s $ and $ g_2^s $ such that $ g^s $ and $ g_1^s $ satisfy
$ g^s (h^{s-1}, x^{s-1}) = g_1^s (h^{s-1}) + g_2^s (x^{s-1}) $ and $ g_1^s(u + v) = g_1^s(u) + g_1^s(v)$ for all $ s $, we have
\begin{align*}
h^t & = g_1^t (h^{t-1}) + g_2^t (x^{t-1}) \\
& = g_1^t (g_1^{t-1} (h^{t-2}) + g_2^{t-1} (x^{t-2})) + g_2^t (x^{t-1}) \\
& = g_1^t (g_1^{t-1} (h^{t-2})) + g_1^t (g_2^{t-1} (x^{t-2})) + g_2^t (x^{t-1}) \\
& = \cdots,
\end{align*}
which is a summation over transformed $ x^{t-1}, x^{t-2}, ..., x^0 $ with a constant $ g_1^t(g_1^{t-1}(\cdots g_1^1(h^0) )) $.

\paragraph{The proposed RLA module in Figure 3}
Consider the RLA module given by Figure 3 with update
\begin{equation*}
h^t=g_2[g_1(y^t)+h^{t-1}]\hspace{5mm}\text{and}\hspace{5mm}
x^t=y^t+x^{t-1},
\end{equation*}
where $y^t=f_1^t [\text{Concat}(h^{t-1}, x^{t-1})]$ and $ g_1, g_2$ are the shared 1x1 and 3x3 convolutions.
As discussed in the main text, ResNets have a layer aggregation interpretation, i.e., $ x^t $ is an aggregation of the residual information $ y^t $.
According to the ablation study in Sections 4.4 and \ref{sec:ablation_cifar}, it is preferable for the RLA module to perform an aggregation of the residual information $ y^t $ instead of the already aggregated $ x^t $.
Thus, in the following, we show that $ h^t $ is an aggregation of $ y^t $ instead of $ x^t $.

When nonlinearities are ignored, the shared convolution $ g_2 $ can be distributed to the two terms, i.e., 
\begin{equation*} 
h^t = g_2(h^{t-1}) + g_2 \circ g_1(y^t),
\end{equation*}
Furthermore, recursively applying the above equation, we have
\begin{align} 
h^t = \, & g_2(h^{t-1}) + g_2 \circ g_1(y^t) \nonumber \\
= \, & g_2(g_2(h^{t-2}) + g_2 \circ g_1(y^{t-1})) + g_2 \circ g_1(y^t) \nonumber \\
= \, & g_2^2(h^{t-2}) + g_2^2 \circ g_1(y^{t-1}) + g_2 \circ g_1(y^t) \nonumber \\
& \vdots \nonumber \\
= \, & \sum_{k=1}^t g_2^k \circ g_1 (y^{t-k+1}) + g_2^t(h^0), \label{eq:RLA-LA}
\end{align}
where $\circ$ denotes the composition of convolution functions, and with a slight abuse of notation, the composition of the same function $k$ times is denoted as its $k$-th power, e.g., 
$$ \begin{matrix} g_2^k = & \underbrace{ g_2 \circ g_2 \circ \cdots \circ g_2 } \\ & k \text{ times} \end{matrix} ,$$
not to be confused with time varying functions where the superscript denotes the time index.
Thus, the RLA hidden feature maps $ h^t $ are aggregations of previous residual information.

Moreover, the patterns of layer aggregation in ResNets and the RLA module in Eq. \eqref{eq:RLA-LA} are very similar to the AR($ \infty $) and ARMA($1, 1$) models introduced in Sections 3.2 and \ref{sec:time-series}.
Because for the proposed RLA module in Figure 3, the convolutions $ g_l^t $ in Eq. (2) are solely determined by the two shared convolutions $ g_2 $ and $g_1$.

\section{Experiments} \label{sec:supp-exp}

Due to limited space in the main paper, we present more experiment results, implementation details and descriptions of network structures in this section.
\begin{revised}
	Our implementation and weights are available at \url{https://github.com/fangyanwen1106/RLANet}.
\end{revised}

\subsection{Implementation details for CIFAR}
The two CIFAR datasets, CIFAR-10 and CIFAR-100, consist of 60k $32 \times 32$ RGB images of 10 and 100 classes, respectively. 
The training and test sets contain 50k and 10k images, and we adopt a train/validation split of 45k/5k.
We follow a standard data preprocessing scheme that is widely used for CIFAR dataset \cite{he2016deep, huang2017densely}. 
Specifically, we normalize the data using the RGB-channel means and standard deviations. 
For data augmentation, the input image is first zero-padded with 4 pixels on each side, resulting in a $40 \times 40$ image. 
A random $32 \times 32$ crop is then performed on the image or its horizontal flip. 

All the networks are trained from scratch using Nesterov SGD with momentum of 0.9, $l_2$ weight decay of $10^{-4}$, and a batch size of 128 for 300 epochs.
The initial learning rate is set to 0.1 and is divided by 10 at 150 and 225 epochs.
We adopt the weight initialization method following \cite{he2015delving}. 
The model with the highest validation accuracy is chosen and reported. 
It is noticeable that we use the same hyperparameters as in \cite{huang2017densely} except that we split validation data and use a batch size of 128.
The original $32 \times 32$ images are used for testing.
All programs run on one Tesla V100 GPU with 32GB memory.


\paragraph{RLA-ResNet-110 for CIFAR}
See Table \ref{tab:archi_res110}. 
In ResNet-110, the bottleneck structure in ResNet-164 is replaced with two 3x3 convolutions.
To control the number of parameters, the number of channels in ResNet-110 is one-fourth of that in ResNet-164.
Following the same spirit, we apply an RLA module with $k = 4$ (instead of 12) for ResNet-110.

\begin{table}[t]
	\caption{Network architectures of ResNet-110 (left) and RLA-ResNet-110 (right). 
		Note that each "conv" layer shown in the table corresponds to the sequence BN-ReLU-Conv, except that the first "conv" layer corresponds to Conv-BN-ReLU.
		RLA channel is specified by the value of k.}
	\label{tab:archi_res110}
	\begin{center}
		\scriptsize
		\begin{tabular}{ccc}
			\toprule
			\multicolumn{1}{l|}{Output size} & \multicolumn{1}{c|}{ResNet-110} & RLA-ResNet-110   \\ 
			\midrule
			\multicolumn{1}{c|}{\multirow{2}{*}{$32 \times 32$}}             & \multicolumn{2}{c}{conv, $3 \times 3$, 16}                \\
			\cline{2-3}           
			\multicolumn{1}{l|}{}  & \multicolumn{1}{c|}{$\left[ \begin{aligned}
				& {\rm conv}, 3\times3, 16 \\
				& {\rm conv}, 3\times3, 16 \\
				\end{aligned} \right] \times 18 $ }  &  \multicolumn{1}{c}{$\left[ \begin{aligned}
				& {\rm conv}, 3\times3, 16 \\
				& {\rm conv}, 3\times3, 16 \\
				& \qquad {\rm k} = 4
				\end{aligned} \right] \times 18 $ }  
			\\  \cline{1-3}
			\multicolumn{1}{c|}{$16 \times 16$}  & \multicolumn{1}{c|}{$\left[ \begin{aligned}
				& {\rm conv}, 3\times3, 32 \\
				& {\rm conv}, 3\times3, 32 \\
				\end{aligned} \right] \times 18 $}  &  \multicolumn{1}{c}{$\left[ \begin{aligned}
				& {\rm conv}, 3\times3, 32 \\
				& {\rm conv}, 3\times3, 32 \\
				& \qquad {\rm k} = 4
				\end{aligned} \right] \times 18 $}               
			\\ \cline{1-3}
			\multicolumn{1}{c|}{$8 \times 8$}  & \multicolumn{1}{c|}{$\left[ \begin{aligned}
				& {\rm conv},  3\times3, 64 \\
				& {\rm conv},  3\times3, 64 \\
				\end{aligned} \right] \times 18 $}  &  \multicolumn{1}{c}{$\left[ \begin{aligned}
				& {\rm conv},  3\times3, 64 \\
				& {\rm conv},  3\times3, 64 \\
				& \qquad {\rm k} = 4
				\end{aligned} \right] \times 18 $}  
			\\  \cline{1-3}
			\multicolumn{1}{c|}{$1 \times 1$}  &  \multicolumn{2}{c}{global average pool, 10-d or 100-d $fc$, softmax}\\
			\bottomrule
		\end{tabular}
	\end{center}
\end{table}

\paragraph{RLA-ResNet-164 for CIFAR}
See Table \ref{tab:archi_res164}.

\begin{table}[t]
	\caption{Network architectures of ResNet-164 (left) and RLA-ResNet-164 (right). 
		Note that each "conv" layer shown in the table corresponds to the sequence BN-ReLU-Conv, except that the first "conv" layer corresponds to Conv-BN-ReLU.
		RLA channel is specified by the value of k.}
	\label{tab:archi_res164}
	\begin{center}
		\scriptsize
		\begin{tabular}{ccc}
			\cline{1-3} 
			\multicolumn{1}{l|}{Output size} & \multicolumn{1}{c|}{ResNet-164} & RLA-ResNet-164   \\ 
			\cline{1-3}
			\multicolumn{1}{c|}{\multirow{2}{*}{$32 \times 32$}}             & \multicolumn{2}{c}{conv, $3 \times 3$, 16}                \\
			\cline{2-3}           
			\multicolumn{1}{l|}{}  & \multicolumn{1}{c|}{$\left[ \begin{aligned}
				& {\rm conv}, 1\times1, 16 \\
				& {\rm conv}, 3\times3, 16 \\
				& {\rm conv}, 1\times1, 64
				\end{aligned} \right] \times 18 $ }  &  \multicolumn{1}{c}{$\left[ \begin{aligned}
				& {\rm conv}, 1\times1, 16 \\
				& {\rm conv}, 3\times3, 16 \\
				& {\rm conv}, 1\times1, 64 \\
				& \qquad {\rm k} = 12
				\end{aligned} \right] \times 18 $ }  
			\\  \cline{1-3}
			\multicolumn{1}{c|}{$16 \times 16$}  & \multicolumn{1}{c|}{$\left[ \begin{aligned}
				& {\rm conv}, 1\times1, 32 \\
				& {\rm conv}, 3\times3, 32 \\
				& {\rm conv}, 1\times1, 128
				\end{aligned} \right] \times 18 $}  &  \multicolumn{1}{c}{$\left[ \begin{aligned}
				& {\rm conv}, 1\times1, 32 \\
				& {\rm conv}, 3\times3, 32 \\
				& {\rm conv}, 1\times1, 128 \\
				& \qquad {\rm k} = 12
				\end{aligned} \right] \times 18 $}               
			\\ \cline{1-3}
			\multicolumn{1}{c|}{$8 \times 8$}  & \multicolumn{1}{c|}{$\left[ \begin{aligned}
				& {\rm conv},  1\times1, 64 \\
				& {\rm conv},  3\times3, 64 \\
				& {\rm conv},  1\times1, 256
				\end{aligned} \right] \times 18 $}  &  \multicolumn{1}{c}{$\left[ \begin{aligned}
				& {\rm conv},  1\times1, 64 \\
				& {\rm conv},  3\times3, 64 \\
				& {\rm conv},  1\times1, 256 \\
				& \qquad {\rm k} = 12
				\end{aligned} \right] \times 18 $}  
			\\  \cline{1-3}
			\multicolumn{1}{c|}{$1 \times 1$}  &  \multicolumn{2}{c}{global average pool, 10-d or 100-d $fc$, softmax}\\
			\cline{1-3}
		\end{tabular}
	\end{center}
\end{table}

\paragraph{RLA-Xception for CIFAR}
Since Xception is not originally implemented for the CIFAR dataset, we modify it into a thinner version with 2.7M parameters by reducing the channels in each layer while keeping the depth unchanged.
We replace the first 7x7 convolution with a 3x3 one and remove three downsampling layers. 
We use average pooling instead of max pooling for downsampling.
For Xception, the architectures have multiple stages with fewer layers in each stage. 
\begin{revised}We group the layers with the same resolutions as a stage like in ResNet.\end{revised}
See Table \ref{tab:archi_xcep_cifar}.

\begin{table*}[t]
	\caption{Network architectures of Xception (left) and RLA-Xception (right) used on CIFAR. 
		Note that all "conv" and "sep conv" (depthwise separable convolution) layers are followed by batch normalization.
		RLA channel is specified by the value of k.}
	\label{tab:archi_xcep_cifar}
	\begin{center}
		\scriptsize
		\begin{tabular}{ccc}
			\toprule
			\multicolumn{1}{l|}{Output size} & \multicolumn{1}{c|}{Xception} & RLA-Xception    \\ 
			\midrule
			\multicolumn{1}{c|}{\multirow{5}{*}{$32 \times 32$}}  & \multicolumn{2}{c}{conv, \ $3 \times 3$, \ 16, \quad relu}      \\ 
			\cline{2-3}
			\multicolumn{1}{l|}{}  & \multicolumn{2}{c}{conv, \ $3 \times 3$, \ 32, \quad relu}                \\
			\cline{2-3}           
			\multicolumn{1}{l|}{}  & \multicolumn{1}{l|}{$\left[ \begin{aligned}
				& {\rm sep \ conv}, 3 \times 3, 64 \\
				& {\rm relu} \\
				& {\rm sep \ conv}, 3 \times 3, 64 
				\end{aligned} \right] $ }  &  \multicolumn{1}{l}{$\left[ \begin{aligned}
				& {\rm sep \ conv}, 3 \times 3, 64 \\
				& {\rm relu} \\
				& {\rm sep \ conv}, 3 \times 3, 64 \\
				& \qquad \quad {\rm k} = 12
				\end{aligned} \right] $ }  
			\\  \cline{2-3}
			
			\multicolumn{1}{l|}{}  & \multicolumn{1}{l|}{$\left[ \begin{aligned}
				& {\rm relu} \\
				& {\rm sep \ conv}, 3 \times 3, 128 \\
				& {\rm relu} \\
				& {\rm sep \ conv}, 3 \times 3, 128 
				\end{aligned} \right] $ }  &  \multicolumn{1}{l}{$\left[ \begin{aligned}
				& {\rm relu} \\
				& {\rm sep \ conv}, 3 \times 3, 128 \\
				& {\rm relu} \\
				& {\rm sep \ conv}, 3 \times 3, 128 \\
				& \qquad \quad {\rm k} = 12
				\end{aligned} \right] $}               
			\\ \cline{2-3}
			
			\multicolumn{1}{l|}{}  & \multicolumn{1}{l|}{$\left[ \begin{aligned}
				& {\rm relu} \\
				& {\rm sep \ conv}, 3 \times 3, 256 \\
				& {\rm relu} \\
				& {\rm sep \ conv}, 3 \times 3, 256 
				\end{aligned} \right] $ }  &  \multicolumn{1}{l}{$\left[ \begin{aligned}
				& {\rm relu} \\
				& {\rm sep \ conv}, 3 \times 3, 256 \\
				& {\rm relu} \\
				& {\rm sep \ conv}, 3 \times 3, 256 \\
				& \qquad \quad {\rm k} = 12
				\end{aligned} \right] $}                
			\\ \cline{1-3}
			
			\multicolumn{1}{c|}{\multirow{2}{*}{$16 \times 16$}}  & \multicolumn{1}{l|}{$\left[ \begin{aligned}
				& {\rm relu} \\
				& {\rm sep \ conv},  3 \times 3, 256 \\
				& {\rm relu} \\
				& {\rm sep \ conv},  3 \times 3, 256 \\
				& {\rm relu} \\
				& {\rm sep \ conv},  3 \times 3, 256
				\end{aligned} \right] \times 8 $}  &  \multicolumn{1}{l}{$\left[ \begin{aligned}
				& {\rm relu} \\
				& {\rm sep \ conv},  3 \times 3, 256 \\
				& {\rm relu} \\
				& {\rm sep \ conv},  3 \times 3, 256 \\
				& {\rm relu} \\
				& {\rm sep \ conv},  3 \times 3, 256 \\
				& \qquad \quad {\rm k} = 12
				\end{aligned} \right] \times 8 $}  
			\\  \cline{2-3}
			
			\multicolumn{1}{l|}{}  & \multicolumn{1}{l|}{$\left[ \begin{aligned}
				& {\rm relu} \\
				& {\rm sep \ conv},  3 \times 3, 256 \\
				& {\rm relu} \\
				& {\rm sep \ conv},  3 \times 3, 512
				\end{aligned} \right] $}  &  \multicolumn{1}{l}{$\left[ \begin{aligned}
				& {\rm relu} \\
				& {\rm sep \ conv},  3 \times 3, 256 \\
				& {\rm relu} \\
				& {\rm sep \ conv},  3 \times 3, 512 \\
				& \qquad \quad {\rm k} = 12
				\end{aligned} \right] $}  
			\\  \cline{1-3}
			
			\multicolumn{1}{c|}{$8 \times 8$}  & \multicolumn{1}{l|}{$\left[ \begin{aligned}
				& {\rm sep \ conv}, 3 \times 3, 512 \\
				& {\rm relu} \\
				& {\rm sep \ conv}, 3 \times 3, 512 \\
				& {\rm relu} \\
				\end{aligned} \right] $}  &   \multicolumn{1}{l}{$\left[ \begin{aligned}
				& {\rm sep \ conv}, 3 \times 3, 512 \\
				& {\rm relu} \\
				& {\rm sep \ conv}, 3 \times 3, 512 \\
				& {\rm relu} \\
				& \qquad \quad {\rm k} = 12
				\end{aligned} \right] $}   \\ 
			\cline{1-3}
			
			\multicolumn{1}{c|}{$1 \times 1$}  &  \multicolumn{2}{c}{global average pool, 10-d or 100-d $fc$, softmax} \\
			\bottomrule
		\end{tabular}
	\end{center}
	\vskip -0.1in
\end{table*}

\subsection{ImageNet Classification} \label{sec:imagenet}

\subsubsection{Implementation details for ImageNet} \label{sec:imagenet-std}
For training RLA-ResNets, we follow the same data augmentation and hyper-parameter settings as in \cite{he2016deep} and \cite{huang2017densely}, which are standard pipelines. 
Specifically, we apply scale augmentation to the original images.
A 224 $\times$ 224 crop is randomly sampled from a scaled image or its horizontal flip. 
Each input image is normalized by RGB-channel means and standard deviations.
All the networks are trained using SGD with momentum of 0.9, $l_2$ weight decay of $10^{-4}$ and a mini-batch size of 256 on 4x V100 GPUs.
We train models for 120 epochs from scratch, and use the weight initialization strategy described in \cite{he2015delving}.
The initial learning rate is set to 0.1 and decreased by a factor of 10 every 30 epochs.
For the light-weighted model MobileNetV2, we trained the model on 2x V100 GPUs within 400 epochs using SGD with weight decay of 4e-5, momentum of 0.9 and mini-batch size of 96, following the settings in \cite{sandler2018mobilenetv2}. 
The initial learning rate is set to 0.045, decreased by a linear decay rate of 0.98.
For evaluation on the validation set, the shorter side of an input image is first resized to 256, and a center crop of $224 \times 224$ is then used for evaluation.

\paragraph{RLA-ResNets for ImageNet}
Table \ref{tab:archi_res50} shows the architectures of RLA-ResNet-50, RLA-ResNet-101 and RLA-ResNet-152. 
According to the experience on CIFAR, we follow the growth rate setting of DenseNet on ImageNet and set $k = 32$ in our RLA module on ImageNet.


\begin{table*}[t]
	\caption{Network architectures of ResNets (left) and RLA-ResNets (right).
		Note that each "conv" layer shown in the table corresponds to the sequence BN-ReLU-Conv, except that the first "conv" layer corresponds to Conv-BN-ReLU.
		Varying the block counts $(B_1, B_2)$ gives rise to ResNet-50 and RLA-ResNet-50 $(B_1 = 4, B_2 = 6)$, ResNet-101 and RLA-ResNet-101 $(B_1 = 4, B_2 = 23)$, and ResNet-152 and RLA-ResNet-152 $(B_1 = 8, B_2 = 36)$.}
	\label{tab:archi_res50}
	\begin{center}
		\scriptsize
		\begin{tabular}{ccc}
			\toprule
			\multicolumn{1}{l|}{Output size} & \multicolumn{1}{c|}{ResNet} & RLA-ResNet   \\ 
			\midrule
			\multicolumn{1}{c|}{$112 \times 112$}            & \multicolumn{2}{c}{conv, $7 \times 7$, 64, stride 2}      \\ 
			\cline{1-3}
			\multicolumn{1}{c|}{\multirow{2}{*}{$56 \times 56$}}             & \multicolumn{2}{c}{max pool, $3 \times 3$, stride 2}                \\
			\cline{2-3}           
			\multicolumn{1}{l|}{}  & \multicolumn{1}{c|}{$\left[ \begin{aligned}
				& {\rm conv}, 1\times1, 64 \\
				& {\rm conv}, 3\times3, 64 \\
				& {\rm conv}, 1\times1, 256
				\end{aligned} \right] \times 3 $ }  &  \multicolumn{1}{c}{$\left[ \begin{aligned}
				& {\rm conv}, 1\times1, 64 \\
				& {\rm conv}, 3\times3, 64 \\
				& {\rm conv}, 1\times1, 256 \\
				& \qquad {\rm k} = 32
				\end{aligned} \right] \times 3 $ }  
			\\  \cline{1-3}
			
			\multicolumn{1}{c|}{$28 \times 28$}  & \multicolumn{1}{c|}{$\left[ \begin{aligned}
				& {\rm conv}, 1\times1, 128 \\
				& {\rm conv}, 3\times3, 128 \\
				& {\rm conv}, 1\times1, 512
				\end{aligned} \right] \times B_{1} $}  &  \multicolumn{1}{c}{$\left[ \begin{aligned}
				& {\rm conv}, 1\times1, 128 \\
				& {\rm conv}, 3\times3, 128 \\
				& {\rm conv}, 1\times1, 512 \\
				& \qquad {\rm k} = 32
				\end{aligned} \right] \times B_{1} $}               
			\\ \cline{1-3}
			
			\multicolumn{1}{c|}{$14 \times 14$}  & \multicolumn{1}{c|}{$\left[ \begin{aligned}
				& {\rm conv},  1\times1, 256 \\
				& {\rm conv},  3\times3, 256 \\
				& {\rm conv},  1\times1, 1024
				\end{aligned} \right] \times B_{2} $}  &  \multicolumn{1}{c}{$\left[ \begin{aligned}
				& {\rm conv},  1\times1, 256 \\
				& {\rm conv},  3\times3, 256 \\
				& {\rm conv},  1\times1, 1024 \\
				& \qquad {\rm k} = 32
				\end{aligned} \right] \times B_{2} $}  
			\\  \cline{1-3}
			
			\multicolumn{1}{c|}{$7 \times 7$}  & \multicolumn{1}{c|}{$\left[ \begin{aligned}
				& {\rm conv}, 1\times1, 512 \\
				& {\rm conv}, 3\times3, 512 \\
				& {\rm conv}, 1\times1, 2048
				\end{aligned} \right] \times 3 $}  &   \multicolumn{1}{c}{$\left[ \begin{aligned}
				& {\rm conv}, 1\times1, 512 \\
				& {\rm conv}, 3\times3, 512 \\
				& {\rm conv}, 1\times1, 2048 \\
				& \qquad {\rm k} = 32
				\end{aligned} \right] \times 3 $}   \\ 
			\cline{1-3}
			
			\multicolumn{1}{c|}{$1 \times 1$}  &  \multicolumn{2}{c}{global average pool, 1000-d $fc$, softmax}\\
			\bottomrule
		\end{tabular}
	\end{center}
	\vskip -0.1in
\end{table*}

\paragraph{RLA-MobileNetV2 for ImageNet}
We follow the guidelines in Section 3.3 in the main paper, and group main CNN layers of MobileNetV2 into 5 stages. 
For each stage, a shared 3x3 depthwise separable convolution is applied in the RLA module.
Different from ResNets, MobileNetV2 uses inverted residual bottleneck block.
To avoid the expansion in model complexity (FLOPs), we make the following modifications according to the properties of the inverted residual block:
\begin{enumerate}
	\item Instead of concatenating the hidden states $h_{t-1}$ and the inputs $x_{t-1}$ before the first 1x1 Conv in the bottleneck block, $h_{t-1}$ and $x_{t-1}$ are concatenated after the first 1x1 Conv to avoid the expansion step in the inverted residual block.
	\item The shared standard convolution is replaced with a shared depthwise separable convolution in the RLA module, which is more compatible with MobileNetV2.
\end{enumerate}

\subsubsection{Experiments with training and architecture refinements} \label{sec:tricks}
\paragraph{Implementation details}
In Sections 4.2 and \ref{sec:imagenet-std}, we adopt the standard training pipeline for fair comparisons and validate the effectiveness of the proposed structure. 
To further improve accuracy, we apply some of training tricks and architecture refinements described in \cite{he2019bag}. 
We apply these training tricks by pytorch-image-models toolkit \cite{rw2019timm}. \footnote{License: Apache License 2.0}
Specifically,
\begin{itemize}
	\item[(a)] we apply label smoothing with $\epsilon = 0.1$ for regularization following \cite{he2019bag};
	\item[(b)] we consider the mixup augmentation method, and we choose $\alpha = 0.2$ in the Beta distribution for mixup training; 
	\item[(c)] we use 5 epochs to gradually warm up learning rate at the beginning of the training; and
	\item[(d)] we use the cosine learning rate schedule within 200 epochs, and set the initial learning rate as 0.1.
\end{itemize}
It is worth pointing out that our training setting above is selected based on training efficiency and does not equal to any setting reported in \cite{he2019bag}. 

We refine the architecture by making the following adjustments to the original ResNets, and the resulting network is often called ResNet-D \cite{he2019bag}. 
\begin{itemize}
	\item[(a)] The stride sizes are switched for the first two convolutions in the residual path of the downsampling blocks. 
	\item[(b)] The 7x7 convolution in the stem is replaced by three smaller 3x3 convolutions. 
	\item[(c)] The stride-2 1x1 convolution in the skip connection path of the downsampling blocks is replaced by stride-2 2x2 average pooling and then a non-strided 1x1 convolution.
\end{itemize}

\subsection{Object detection and instance segmentation on MS COCO}
\paragraph{Implementation details} 
To show the transferability and the generalization ability, we experiment our RLA-Nets on the object detection task \begin{revised}using three typical object detection frameworks: Faster R-CNN \cite{fasterrcnn2015}, Mask R-CNN \cite{he2017mask} and RetinaNet \cite{lin2017focal}.\end{revised}
For Mask R-CNN, we also show instance segmentation results.
ResNet-50 and ResNet-101, pretrained on ImageNet, along with FPN \cite{lin2017fpn} are used as backbone models. 
All detectors are implemented by open source MMDetection toolkit \cite{mmdetection}. \footnote{License: Apache License 2.0}
We employ the default settings to finetune our RLA-Nets on COCO train2017 set, and evaluate the performance on COCO val2017 set.
Specifically, the shorter side of input images are resized to 800. 
We train all detectors within 12 epochs using SGD with weight decay of 1e-4, momentum of 0.9 and mini-batch size of 8. 
The learning rate is initialized to 0.01 and is decreased by a factor of 10 after 8 and 11 epochs, respectively, i.e., the 1x training schedule. 

The results shown in Tables 5 and 6 are all obtained from the models pretrained with standard training pipelines on ImageNet.
The models with our proposed RLA modules show strong advantages in transfer learning, which significantly improve both object detection and instance segmentation.
Thus, we make a reasonable conjecture that the performances of these tasks could be further improved with the help of the pretrained models with training and architecture refinements on ImageNet, as the transfer learning results in \cite{he2019bag}.

\begin{revised}
	\subsection{Ablation study} \label{sec:ablation_cifar}
	
	In this subsection, we report ablation experiments on CIFAR with ResNet-164 as the baseline network. 
	As in the ablation study on ImageNet in the main paper, we compare RLA modules with the same set of variations.
	Besides, we also compare with its DenseNet variant, i.e., dual path network (DPN) \cite{chen2017dual}.
	And different values of RLA channel $k$ are examined as well.
	
	Moreover, we present additional results and discussions on the parallelizability, throughput and weight-sharing of RLA-Net in Sections \ref{sec:time} and \ref{sec:share}.
	
	\begin{table}[ht]
		\caption{Classification errors on the CIFAR-10 test set using ResNet-164 as the baseline model.}
		\label{tab:ablation-structure}
		\centering
		\small
		\begin{tabular}{lcc}
			\toprule
			Model  &  Params  &  Error (\%)  \\
			\midrule
			ResNet-164 &  1.72M  &  5.72 $\pm$ 0.02  \\
			- channel +12  &  1.93M  &  5.27  \\
			- RLA-v1 ($k=12$, unshared)  &  1.90M  &  5.17  \\
			- RLA-v1 ($k=12$, no exchange) &  1.72M  &  5.51  \\
			- RLA-v1 ($k=12$, ConvLSTM)  &  1.77M  &  5.36  \\
			- RLA-v1 ($k=12$, PostAct.)  &  1.74M  &  5.39  \\
			- DenseNet ($k=12$, DPN) &  2.13M  &  5.99  \\
			\midrule
			- \bf{RLA-v1 ($k=12$)}   &  1.74M  &  \bf{4.95}  \\
			- RLA-v2 ($k=12$)   &  1.74M  &  5.53  \\
			- RLA-v3 ($k=12$)   &  1.74M  & 5.13   \\
			- RLA-v4 ($k=12$) 	&  1.74M  &  5.35  \\
			- RLA-v5 ($k=12$)	&  1.74M  &  5.00  \\
			- RLA-v6 ($k=12$)	&  1.74M  &  5.62  \\
			\bottomrule
		\end{tabular}
	\end{table}

	\subsubsection{Comparison of RLA module and its variants}
	The results are reported in Table \ref{tab:ablation-structure}.
	Again, we find that sharing the convolutions are indeed effective, and the improvement in accuracy is not fully caused by the increment in the network width.
	As DPN is not originally implemented for CIFAR and ResNet-164, we modified ResNet-164 following \cite{chen2017dual}.
	To our surprise, our implementation of DPN does not perform well in this experiment.
	For other variants, we also obtain consistent observations with the conclusions in the main paper.
	Detailed network architectures are introduced below.
	
	
	\paragraph{ResNet-164 (channel+12)} 
	For easy comparison, we repeat Table \ref{tab:archi_res164} here with an additional column describing the structure of ResNet-164 (channel+12); see Table \ref{tab:archi_res164abl}.
	
	\begin{table*}[t]
		\caption{Network architectures of ResNet-164 (left), ResNet-164 (channel+12) (middle) and RLA-ResNet-164 (right). 
			Note that each "conv" layer shown in the table corresponds to the sequence BN-ReLU-Conv, except that the first "conv" layer corresponds to Conv-BN-ReLU.}
		\label{tab:archi_res164abl}
		\begin{center}
			\scriptsize
			\begin{tabular}{cccc}
				\toprule
				\multicolumn{1}{l|}{Output size} & \multicolumn{1}{c|}{ResNet-164} & \multicolumn{1}{c|}{ResNet-164 (channel+12)} & RLA-ResNet-164   \\ 
				\midrule
				\multicolumn{1}{c|}{\multirow{2}{*}{$32 \times 32$}} & \multicolumn{3}{c}{conv, $3 \times 3$, 16}                \\
				\cline{2-4}           
				\multicolumn{1}{l|}{}  
				& \multicolumn{1}{c|}{$\left[ \begin{aligned}
					& {\rm conv}, 1\times1, 16 \\
					& {\rm conv}, 3\times3, 16 \\
					& {\rm conv}, 1\times1, 64
					\end{aligned} \right] \times 18 $ }  
				& \multicolumn{1}{c|}{$\left[ \begin{aligned}
					& {\rm conv}, 1\times1, 16+3 \\
					& {\rm conv}, 3\times3, 16+3 \\
					& {\rm conv}, 1\times1, 64+12
					\end{aligned} \right] \times 18 $ }  
				&  \multicolumn{1}{c}{$\left[ \begin{aligned}
					& {\rm conv}, 1\times1, 16 \\
					& {\rm conv}, 3\times3, 16 \\
					& {\rm conv}, 1\times1, 64 \\
					& \qquad {\rm k} = 12
					\end{aligned} \right] \times 18 $ }  
				\\  \cline{1-4}
				
				\multicolumn{1}{c|}{$16 \times 16$}  
				& \multicolumn{1}{c|}{$\left[ \begin{aligned}
					& {\rm conv}, 1\times1, 32 \\
					& {\rm conv}, 3\times3, 32 \\
					& {\rm conv}, 1\times1, 128
					\end{aligned} \right] \times 18 $}  
				& \multicolumn{1}{c|}{$\left[ \begin{aligned}
					& {\rm conv}, 1\times1, 32+3 \\
					& {\rm conv}, 3\times3, 32+3 \\
					& {\rm conv}, 1\times1, 128+12
					\end{aligned} \right] \times 18 $}  
				&  \multicolumn{1}{c}{$\left[ \begin{aligned}
					& {\rm conv}, 1\times1, 32 \\
					& {\rm conv}, 3\times3, 32 \\
					& {\rm conv}, 1\times1, 128 \\
					& \qquad {\rm k} = 12
					\end{aligned} \right] \times 18 $}               
				\\ \cline{1-4}
				
				\multicolumn{1}{c|}{$8 \times 8$}  
				& \multicolumn{1}{c|}{$\left[ \begin{aligned}
					& {\rm conv},  1\times1, 64 \\
					& {\rm conv},  3\times3, 64 \\
					& {\rm conv},  1\times1, 256
					\end{aligned} \right] \times 18 $}  
				& \multicolumn{1}{c|}{$\left[ \begin{aligned}
					& {\rm conv},  1\times1, 64+3 \\
					& {\rm conv},  3\times3, 64+3 \\
					& {\rm conv},  1\times1, 256+12
					\end{aligned} \right] \times 18 $}  
				&  \multicolumn{1}{c}{$\left[ \begin{aligned}
					& {\rm conv},  1\times1, 64 \\
					& {\rm conv},  3\times3, 64 \\
					& {\rm conv},  1\times1, 256 \\
					& \qquad {\rm k} = 12
					\end{aligned} \right] \times 18 $}  
				\\  \cline{1-4}
				
				\multicolumn{1}{c|}{$1 \times 1$}  &  \multicolumn{3}{c}{global average pool, 10-d or 100-d $fc$, softmax}\\
				\bottomrule
			\end{tabular}
		\end{center}
		\vskip -0.2in
	\end{table*}

	\paragraph{ConvLSTM based RLA} 
	A diagram for ConvLSTM based RLA module is shown in Figure \ref{fig:RLA-lstm}.
	Different from \cite{shi2015convolutional}, our ConvLSTM cell is the convolutional counterpart of the conventional LSTM cell, instead of the fully connected LSTM (FC-LSTM) cell.
	All the cell updates are exactly the same as those in the Keras official implementation of the class \texttt{ConvLSTM2DCell}.
	
	\begin{figure}[ht]
		\begin{center}
			\centerline{\includegraphics[width=0.3\textwidth]{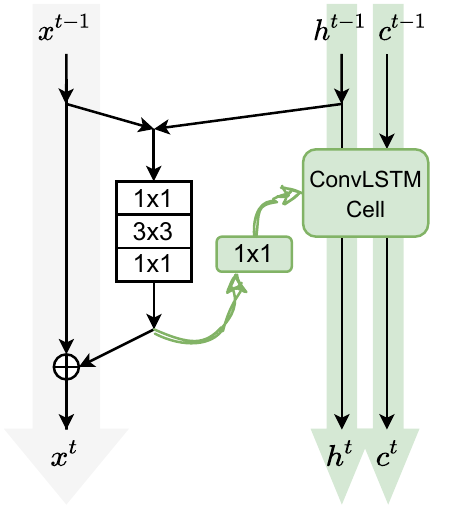}}
			\caption{Diagram of the ConvLSTM variant of the recurrent layer aggregation (RLA) module.}
			\label{fig:RLA-lstm}
		\end{center}
		\vskip -0.2in
	\end{figure}
	
	\paragraph{Post-Activated RLA} 
	Motivated by \cite{he2016identity}, we experimented with the post-activated RLA module, see Figure \ref{fig:RLA-post}.
	Compared with Figure 3 in the main text, we only change the sequence BN-tanh-Conv to Conv-BN-tanh.
	It turns out that pre-activated RLA performs better on the pre-activated ResNet-164 model on CIFAR-10.
	In case of post-activated CNN-based models, post-activated RLA may still be worth trying.
	
	\begin{figure}[ht]
		\begin{center}
			\centerline{\includegraphics[width=0.35\textwidth]{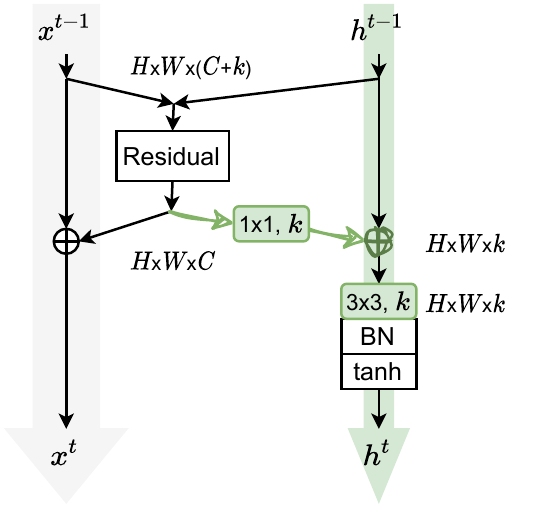}}
			\caption{Diagram of the post-activated variant of the recurrent layer aggregation (RLA) module.}
			\label{fig:RLA-post}
		\end{center}
		\vskip -0.2in
	\end{figure}
	

	\subsubsection{Comparison of integration strategies}
	We compare different strategies to combine our RLA module with the main CNN.
	In addition to the proposed design, we consider five variants (v2--v6), as depicted in Figure 4 in the main text.
	The performance of the variants are reported in Table \ref{tab:ablation-structure}.
	The results on CIFAR are consistent with those on ImageNet, further validating our perspectives and statements in the main text.
	

	\subsubsection{RLA channel $k$}
	RLA channel $k$ is defined as the number of filters in the shared 3x3 convolution in the RLA module. 
	This hyperparameter allows us to vary the capacity and computational cost of the RLA module in a network. 
	Based on RLA-v1, we experiment with various $k$ values in the set \{8, 12, 16, 24\}.
	Table \ref{tab:ablation-k} shows that the performance is robust to different $k$ values, and the increase in complexity does not improve accuracy monotonically.
	Setting $k = 12$ strikes a good balance between accuracy and complexity.
	This result is consistent with the growth rate setting of DenseNet.  
	
	\begin{table}[ht]
		\caption{Classification error on CIFAR-10 using ResNet-164 with different RLA channel $k$.}
		\label{tab:ablation-k}
		\begin{center}
			\begin{small}
				\begin{tabular}{lcc}
					\toprule
					Model  &  Params  &  Error (\%)  \\
					\midrule
					RLA-v1 ($k=8$)  &  1.73M  &  4.98  \\
					\bf{RLA-v1 ($k=12$)}  &  1.74M  &  \bf{4.95}  \\
					RLA-v1 ($k=16$)  &  1.75M  &  5.28  \\
					RLA-v1 ($k=24$)  &  1.78M  &  5.14  \\
					\bottomrule
				\end{tabular}
			\end{small}
		\end{center}
	\end{table}
	
\end{revised}


\begin{revised}
	
	\subsubsection{Parallelizability and throughput of RLA-Net} \label{sec:time}
	
	Thanks to an anonymous reviewer's comment, we discuss the parallelizability and throughput of RLA-Net in this subsection.
	It is noticeable that adding RLA modules does not affect the parallelizability of the CNN because all existing parallelizable dimensions (batch, channel, width, height) are still parallelizable.
	Not like the commonly used RNNs in the literature, the proposed RLA module treats the layers of a deep CNN as inputs, while the feedforward procedure of a CNN is not parallelizable with respect to its depth. 
	Thus, adding RLA does not affect the parallelizability of the resulting network, and the increment of complexity is similar to adding CNN modules, like SE- or ECA-block.
	
	The throughput is measured as the number of images processed per second on a GPU.
	We present in Table \ref{tab:time} some figures on the time cost of the proposed RLA module in the training and evaluation processes. 
	Specifically, the evaluation is conducted on ImageNet validation set on 1x A100 GPU. 
	Besides, we provide the training speed which has been conducted on 4x V100 GPUs previously.
	From the table below, it can be seen that adding SE, ECA or RLA module to ResNet50 costs about 22\%, 21\% or 26\% more training time. 
	Compared with the original ResNet50, introducing SE, ECA or RLA module costs about 0.2\%, 0.9\% and 2.8\% more time when evaluating on ImageNet validation set.
	
	\begin{table}[ht]
		\caption{Training and evaluation time of different modules using ResNet-50 as the backbone model.}
		\label{tab:time}
		\centering
		\small
		\begin{tabular}{lcc}
			\toprule
			Model  &  Train (s/epoch)  &  Evaluation (ms/image)  \\
			\midrule
			ResNet-50   &   961  &  1.16  \\
			+SE  		&  1171  &  1.17  \\
			+ECA  		&  1161  &  1.17  \\
			+RLA (Ours) &  1211  &  1.20  \\
			\bottomrule
		\end{tabular}
	\end{table}
	
	\subsubsection{Discussions on weight-sharing} \label{sec:share}
	Thanks to an anonymous reviewer's comment, we discuss why we propose weight-sharing in RLA in this subsection.
	Our proposed concept of layer aggregation contains a broad class of implementations including not only shared RLA but also its unshared version. 
	Compared with shared RLA, the unshared version is more general but has more parameters.
	
	We first explain why shared weights, i.e., the shared 1x1 and 3x3 Conv in Figure 3, are used at each layer of the RLA module.
	Weight-sharing is an important feature of the proposed RLA module, and the RLA path actually will become a deep CNN if unshared weights are used.
	While parameter sharing limits the expressiveness of a network, in practice, we have many examples where adding constraints helps. 
	For example, 
	convolution, as a regularized operation, incorporates our prior belief/inductive bias that contents in an image should be transition invariant.
	Similarly, weight-sharing in our RLA module represents our inductive bias that more distant layers could be less dependent; see Figure 2 (left) for support. 
	Besides, it forces the module to learn a pattern different from the main CNN, i.e., ResNet corresponds to the shared-ordinal pattern (see Eq. 10), and RLA corresponds to a different shared-lag pattern (see Eq. 9). 
	If the module is unshared, both the main CNN and the module learn the same shared-ordinal pattern, which possibly leads to redundancy and deteriorates the performance. 
	Thus, we hypothesize that parameter sharing forces the RLA module to take up a functionality different from the main CNN, which turns out to be more helpful for the overall performance.
	
	This inductive bias is further supported by our ablation experiments in Section 4.4. 
	We repeat the results of these experiments in Table \ref{tab:unshared}.
	To provide more experiment results to support our heuristic, we also conduct experiments using ResNet-101 on ImageNet. 
	Specifically, we have spotted advantages of our proposed RLA over its unshared variant in terms of both accuracy (+0.22\%) and parameter count (-10\%) on the ResNet-164 backbone on the CIFAR-10 dataset.
	Similar phenomena can be observed on ImageNet, where our proposed RLA module achieves +0.10\% and +0.28\%  higher accuracy with 2\% and 3\% fewer parameters, respectively. 
	Compared with ResNet-50, the differences on ResNet-101 are more significant. 
	This can be explained that the 3rd stage of ResNet-101 is much longer, leading to the larger differences between shared and unshared versions.
	Specifically, for ResNet-101, the 3rd stage has 23 residual blocks, while it only has 6 residual blocks for ResNet-50.
	These consistent results in Table \ref{tab:unshared} show that the inductive bias embedded in parameter-sharing is beneficial in terms of the trained model. 
	
	\begin{table}[ht]
		\caption{Comparisons of shared RLA and its unshared version.}
		\label{tab:unshared}
		\centering
		\small
		\begin{tabular}{llcccc}
			\toprule
			Data & Model  &  Params  &  FLOPs & Top-1 acc. & Top-5 acc. \\
			\midrule
			\multirow{2}{*}{CIFAR-10} & RLA-ResNet-164 (Ours) & 1.74M & 8.74M & 95.05 & - \\
			& -Unshared variant & 1.90M & 8.74M & 94.83 & - \\
			\midrule
			\multirow{4}{*}{ImageNet} & RLA-ResNet-50 (Ours) & 24.67M & 4.17G & 77.17 & 93.42 \\
			& -Unshared variant & 25.12M & 4.17G & 77.07 & 93.39 \\
			& RLA-ResNet-101 (Ours) & 42.92M & 7.79G & 78.52 & 94.20 \\
			& -Unshared variant & 44.05M & 7.79G & 78.24 & 93.97 \\
			\bottomrule
		\end{tabular}
	\end{table}

	To recommend a specific architecture, we aim at providing an "optimized" structure based on our experiments. 
	Our results show that there is no downside to using the recurrent version. That's why we mainly propose RLA.
	

\end{revised}

\end{document}